\documentclass[pdflatex,sn-mathphys-num]{sn-jnl}
\usepackage{graphicx}
\usepackage{amsmath,amssymb,amsfonts}
\usepackage{amsthm}
\usepackage{mathrsfs}
\usepackage[title]{appendix}
\usepackage{xcolor}
\usepackage{textcomp}
\usepackage{manyfoot}
\usepackage{algorithm}
\usepackage{algorithmicx}
\usepackage{algpseudocode}
\usepackage{listings}
\theoremstyle{thmstyleone}
\newtheorem{theorem}{Theorem}

\theoremstyle{thmstyletwo}

\theoremstyle{thmstylethree}

\usepackage[T1]{fontenc}
\usepackage{subcaption}
\usepackage{natbib}
\usepackage{caption}
\usepackage{multirow}
\usepackage{booktabs,longtable}
\usepackage{tikz}
\usetikzlibrary{shapes.geometric, arrows}
\usepackage{tabularx}
\usepackage{float}
\usepackage[utf8]{inputenc}

\newtheoremstyle{remarkstyle}
  {6pt}
  {6pt}
  {\normalfont}
  {}
  {\bfseries}
  {.}
  { }
  {}

\usepackage{amsthm}
\theoremstyle{remarkstyle}
\newtheorem{remark}{Remark}
\begin{document}

\title[CG4AI: A Column Generation Framework for Training AI Models Under Hard Constraints]{CG4AI: A Column Generation Framework for Training AI Models Under Constraints}

\author[]{\fnm{Youcef} \sur{Magnouche}}
\author[]{\fnm{Abderrahmane} \sur{Driouch}}\email{driouch.ab.dr@gmail.com}
\author[]{\fnm{Sébastien} \sur{Martin}}
\author[]{\fnm{Pierre} \sur{Bauguion}}
\email{firstname.lastname@huawei.com}

\affil[]{\orgdiv{Huawei Technologies Ltd.},
         \orgname{France Research Center},
         \orgaddress{\street{18 Quai du Point du Jour},
                     \city{Boulogne-Billancourt},
                     \postcode{92100},
                     \country{France}}}

\abstract{
Standard machine-learning training minimizes a loss function over a
dataset, but does not guarantee that the resulting model will satisfy
predefined rules or constraints on its outputs. In many real-world
applications, ranging from autonomous systems to network routing, such
guarantees are essential. We propose \emph{CG4AI}, a framework that
builds a convex combination of AI models while enforcing linear
constraints on the combined output. A master linear program (LP)
determines the optimal mixture weights, while a pricing subproblem
generates new models guided by LP dual variables, focusing attention on
the most violated constraints. A cutting-plane procedure extends
feasibility guarantees beyond the training set. We apply CG4AI to two
problems: (i) digit classification on MNIST, where we demonstrate four
distinct uses of constraints, learning from constraints alone,
improving adversarial robustness, correcting misclassified examples, and enforcing output relabeling; and (ii) the multi-commodity flow problem, where link capacity constraints are enforced on neural-network routing
predictors. Experiments on MNIST and standard SNDLIB
benchmark networks show that CG4AI reliably produces feasible predictors
while achieving better accuracy than single-model baselines.
}


\maketitle

\section{Introduction}\label{sec:intro}

Artificial intelligence models are increasingly used to support or
automate decisions in critical domains such as transportation,
healthcare, and telecommunications. These models are typically trained
by minimizing a loss function over a labeled dataset. This process does
not provide any explicit guarantee that predictions will respect safety
rules, physical laws, or regulatory requirements: a model can behave
correctly on the training data and still produce infeasible or unsafe
outputs on new inputs.

This limitation becomes particularly important when predictions must
satisfy \emph{hard constraints}, strict conditions that cannot be
violated under any circumstances. Hard constraints arise naturally in
many settings, as for autonomous vehicles to enforce a collision-avoidance rule, or on radiotherapy planning where the dose delivered to healthy tissue must not exceed a given bound for any tumor configuration. They also arise on network routing to predict as a split of traffic satisfying link capacity constraints. Another interesting application is in regulation and compliance where a lending model must never violate legal fairness requirements, even for unusual applicant profiles. 

Enforcing such constraints in learned models is challenging. The most
common approach adds penalty terms to the training objective to
discourage violations. This transforms hard constraints into soft ones
and cannot guarantee feasibility at test time. Tuning penalty weights is
also notoriously difficult. More recent methods project the model output
onto a feasible set~\citep{min2024hardnet,tordesillas2023rayen} or
embed convex optimization as a differentiable
layer~\citep{amos2017optnet}. Projection methods enforce constraints
exactly but require solving an optimization problem at inference time,
which increases latency and limits scalability. Optimization layers work
well for convex constraints but do not extend naturally to
combinatorial or large-scale settings.

In this paper we propose CG4AI, a framework that enforces hard linear
constraints \emph{during training} using column generation. Instead of
training a single model, we build a convex ensemble: the final predictor
is a weighted combination of simpler models (the \emph{columns}), and
the weights are determined by a master LP. Constraints on the output of
the ensemble enter the master LP directly as linear inequalities, which
ensures that the combined predictor is feasible whenever the LP is
feasible. Each new column is obtained by solving a \emph{pricing
problem} that minimizes the reduced cost of the LP: a quantity combining
the supervised training loss with the dual values of the most violated
constraints. The dual variables act as importance weights during
training, directing the new model toward the inputs where the current
ensemble fails to satisfy the constraints. When the set of constraints is large, a cutting-plane procedure adds only the
most violated constraints at each iteration, keeping the LP tractable.

The key advantage of CG4AI over penalty-based and projection-based
approaches is that constraint satisfaction is enforced at the level of
the LP, not through soft penalties or post-hoc corrections. Once the LP
is feasible, the convex combination of models is guaranteed to satisfy
all enforced constraints, independently of the architecture of the
individual models. The CG4AI considers any kind of AI model as sub-problem while it accepts a custom loss function. 

\subsection*{Contributions}

The proposed mechanism of adding output satisfaction constraints to a
column-generation LP allows us to make the following contributions.
\begin{enumerate}
\item \textbf{CG4AI: LP-certified ensemble learning under output constraints.}
  We propose a general framework that builds a convex ensemble of AI
  models whose combined output satisfies a given set of linear constraints.
  A master LP (with the same column-generation structure as
  LPBoost~\citep{demiriz2002lpboost}, but a different objective and a
  new family of output constraints) determines the mixture weights; a
  pricing subproblem trains new models guided by LP dual variables.
  This extends LPBoost in three ways:
  (a)~the master LP contains \emph{external} constraints on the combined outputs, encoding safety, physical, or user-defined requirements;  (b)~the constraints are associated with AI models inputs that can be entirely disjoint from the training data, separating what the model \emph{must} satisfy from what it \emph{optimizes};
  
  (c)~multiple constraint types are naturally handled, including
  adversarial robustness, correction of misclassified examples, and
  output relabeling, without modifying the framework.

\item \textbf{LP-based training vs.\ inference-time projection.}
  Unlike HardNet~\citep{min2024hardnet} and RAYEN~\citep{tordesillas2023rayen},
  which project model outputs onto a feasible set at every forward pass,
  CG4AI enforces constraints during training through the LP master.
  Once built, the Convex Ensemble of AI models (CE)  inherits feasibility from the LP with zero
  runtime overhead at inference.

\item \textbf{Cutting-plane extension to continuous input spaces.}
  A separation oracle searches for the most violated input in
   the constraint data set and adds it as a new LP row, extending the feasibility
  guarantee beyond any finite constraint set.

\item \textbf{Two applications.}
  On MNIST (proof of concept), we illustrate four constraint types:
  learning from constraints alone, adversarial robustness, misclassification
  correction, and output relabeling. On the multi-commodity flow problem
  (main use case), we enforce link capacity constraints on neural-network
  routing predictors across 41 SNDLIB benchmark configurations.
\end{enumerate}

The paper is organized as follows. Section~\ref{sec:related} reviews
related work. Section~\ref{sec:framework} describes the CG4AI framework.
Sections~\ref{sec:mnist} and~\ref{sec:mcf} present the two
applications. Section~\ref{sec:conclusion} provides conclusions and new research directions.

\section{Related Work}\label{sec:related}

We review five lines of work that are directly relevant to CG4AI:
penalty and Lagrangian methods, differentiable optimization layers,
projection and architectural approaches, MIP-based formulations, and
the use of column generation in machine learning.

\paragraph{Penalty and Lagrangian methods.}
The most common approach for incorporating constraints in learning adds
penalty terms for violations to the training loss. While easy to
implement, these soft-constraint methods cannot guarantee feasibility at
inference time and require careful tuning of penalty
weights~\citep{goodfellow2016deep}. To provide a more principled
treatment, \citet{chamon2020pac} developed a generalization theory for
constrained learning based on the PAC framework, showing that any
PAC-learnable class is also learnable under additional constraints, using
dual-ascent algorithms. \citet{elenter2024near} extended this work and
showed that primal iterates of dual-ascent methods can be near-feasible
and near-optimal in over-parameterized regimes. Augmented Lagrangian
methods have been applied to physics-informed neural networks to enforce
hard equality constraints~\citep{lu2021hpinns}, but they do not
naturally generalize to inequality or combinatorial constraints.

\paragraph{Differentiable optimization layers.}
\citet{amos2017optnet} introduced OptNet, which embeds a convex
quadratic program as a differentiable layer within a neural network,
allowing constraint-aware end-to-end training. Feasibility is enforced
for inputs seen during the forward pass, but not for arbitrary unseen
inputs, and the approach becomes costly at scale.
\citet{agrawal2019differentiable} generalized this to arbitrary convex
cone programs through the CVXPYlayers framework.
\citet{donti2021dc3} proposed DC3, which enforces hard constraints via a
differentiable completion-and-correction procedure: equality constraints
are satisfied by implicit completion, while inequality constraints are
handled by unrolled gradient steps. DC3 does not guarantee feasibility
in general and is sensitive to the number of gradient steps.

\paragraph{Projection and architectural methods.}
\citet{min2024hardnet} proposed HardNet, a projection-based framework
that guarantees constraint satisfaction by projecting network outputs
onto the feasible region after each forward pass, with universal
approximation guarantees. \citet{tordesillas2023rayen} introduced RAYEN,
which parametrizes a feasible convex region analytically and projects
efficiently onto it without iterative solvers, focusing on continuous
convex constraints. \citet{balestriero2023police} proposed POLICE, which
enforces affine output constraints by construction through
a reformulation of the network as a continuous piecewise affine mapping
over a given region; this avoids any runtime overhead but is limited to
a single convex region and affine constraints.
\citet{constante2025enforcing} combine a task network (minimizing loss)
and a safe network (satisfying constraints) through a fixed convex
combination, where feasibility of the combination is guaranteed by
robust optimization duality. All these projection and architectural
methods are limited to convex or affine constraints and do not scale
naturally to combinatorial or high-dimensional settings.

\paragraph{MIP-based formulations.}
\citet{aftabi2024ffnnmip} formulate trained neural networks with binary
activations as mixed-integer programs, enabling exact enforcement of
combinatorial constraints. The approach is theoretically complete but
does not scale to large networks. Related work by \citet{tjeng2019evaluating} uses MIP formulations
to evaluate the robustness of neural networks, and
\citet{anderson2020strong} develop strong relaxations for ReLU networks
in the MIP framework.

\paragraph{Column generation in machine learning.}
Column generation has been used in machine learning primarily in the
context of boosting. \citet{demiriz2002lpboost} proposed LPBoost, which
casts the boosting problem as a linear program and solves it via column
generation: the pricing problem generates new weak learners guided by
dual misclassification costs. \citet{aziz2024ensemble} extended this
idea to deep CNN base learners on CIFAR-10.

\paragraph{CG4AI as an extension of LPBoost.}
CG4AI builds directly on the LPBoost framework
of~\citet{demiriz2002lpboost}. Let $n\in \mathbb N$ be the size of the training data set, and $I$ the set of all possible AI models.  LPBoost solves the following LP over
mixture weights $\lambda_i \geq 0$ for learners $s_i : \mathbb R^d\to\{-1,+1\}$:  
\begin{equation}\label{eq:lpboost}
\max \rho - \nu \sum_{j=1}^n \delta_j
\quad\text{s.t.}\quad \sum_{i\in I} \lambda_i = 1,
\sum_{i\in I} \lambda_i\, s_i(\bar x_j)\, \hat{y}_j + \delta_j \;\geq\; \rho ,\
\forall j=\overline{1,n}
\end{equation} 
where $\bar x_j\in \mathbb R^d$ is the $j$-th training example, $\hat{y}_j \in \{-1,+1\}$
its label, $\rho$ the minimum margin to be maximized, $\delta_j \geq 0$ the
per-example slack, and $\nu \in \mathbb R^+$ a regularization parameter. The
dual variables of the margin constraints play the role of misclassification weights (analogous to AdaBoost), and the column-generation pricing finds the learner $s_i$ that maximally reduces the weighted training error.

CG4AI retains the column-generation structure of LPBoost
(column index set $I$, mixture weights $\lambda_i$, convexity constraint), where $s_i$ is a general AI model rather than a binary classifier i.e., $s_i:\mathbb R^d \Rightarrow \mathbb R^r$. It differs in two fundamental ways. \emph{(i)~Different objective.} CG4AI minimizes
a training loss $\sum_i \mathcal{L}_i \lambda_i$ rather than maximizing a
margin; this allows any differentiable loss (cross-entropy, MSE, etc.)
and is not restricted to binary classification. \emph{(ii)~New family
of output constraints.} In addition to the convexity constraint
$\sum_i \lambda_i = 1$, CG4AI adds rows of the form $\sum_{i \in I} \lambda_i \sum_{q = 1}^ra^q_j s^q_i(\bar{x}_j) \geq\; b_j$
for each point $j$ in a constraint set that can be entirely disjoint from
the training set.
These constraints encode external requirements on the model's output
(safety rules, physical limits, user-defined requirements) and are
formally defined in Section~\ref{sub:master}.
The dual variable of a row $j$ measures how difficult
it is to satisfy the constraint at $\bar{x}_j$ and guide the pricing
toward the most violated inputs.

The nature of the guarantee changes accordingly: LPBoost provides a
probabilistic generalization bound via large training margins. CG4AI
provides a \emph{deterministic feasibility guarantee}: when the LP is
feasible, the ensemble satisfies all output constraints, regardless of whether those inputs appeared during training.
The empirical evidence for this gain is reported in Section~\ref{sub:mnist-comparison}.

\paragraph{Distinction from HardNet and projection methods.}
\citet{min2024hardnet} (HardNet) and~\citet{tordesillas2023rayen} (RAYEN)
guarantee constraint satisfaction by \emph{projecting} the model's output
onto the feasible set at every forward pass, during both training and
inference. This projection happens after the model computes its output,
which means (i) a dedicated optimization problem must be solved at
inference time for every new input, adding computational overhead; (ii)
the projection is defined only over a fixed convex or affine feasible
region and does not naturally extend to combinatorial or input-dependent
constraints; and (iii) the model is not directly trained to satisfy
constraints, it is trained to minimize a loss and then corrected. In
CG4AI, constraints are enforced in the LP master, which directly
determines the mixture weights that make the ensemble feasible. No
projection is needed at inference: the ensemble $f = \sum_i \lambda_i s_i$
inherits feasibility from the LP, and each column $s_i$ is trained by the
pricing to \emph{actively contribute} to constraint satisfaction, guided
by the dual variables. CG4AI is therefore a training-time rather than
an inference-time method, with no runtime overhead after the ensemble is
built.

\medskip
CG4AI's three distinguishing properties are thus: (1) hard constraints
are enforced in the LP master and hold for any convex combination of the
columns; (2) the pricing trains new models guided by LP dual variables,
directing capacity toward the most violated constraints; and (3) the
constraint set can differ from the training set, allowing
user-defined requirements to be imposed without retraining from scratch.

\section{The CG4AI Framework}\label{sec:framework}

We consider the following general setting. In contrast with LPBoost the size of training data and the number of hard constraints may be different. Let $n\in \mathbb N$ be the size of the training data set and $m \in \mathbb N$ the size of the hard constraints set. Let
$\mathcal{I}^{Train} = \{(\bar{x}_k, \hat{y}_k)\}_{k=1}^n$ be a
training dataset, where $\bar{x}_k \in \mathbb{R}^d$ is an input and
$\hat{y}_k \in \mathbb{R}^{r}$ is the corresponding label. Let
$\mathcal{I}^{Const} = \{\bar{x}_j\}_{j=1}^m$ be a set of inputs on which
output constraints must hold; $\mathcal{I}^{Const}$ may overlap with
$\mathcal{I}^{Train}$ or may be disjoint. The goal is to find a
linear combination of predictors $f : \mathbb{R}^d \to \mathbb{R}^r$ that solves:

\begin{equation}\label{eq:clearn}
\min_{f} \;\sum_{k=1}^n L\bigl(f(\bar{x}_k), \hat{y}_k\bigr)
\quad\text{s.t.}\quad
\sum_{q = 1}^ra^q_j f_q(\bar{x}_j) \;\geq\; b_j, \quad \forall j = 1, \ldots, m,
\end{equation}

where $a^q_j \in \mathbb{R}$ is the coefficient associated to $q^{th}$ output of $f$, $b\in \mathbb{R}^m$, and $L$ is a
loss function (e.g., cross-entropy or mean squared error). This problem
is generally intractable for expressive model classes such as deep neural
networks because the constraints impose nonlinear requirements on the
model parameters.

\subsection{Convex Ensemble Structure}\label{sub:ensemble}

Let $\mathcal{M}$ be a large (potentially infinite) set of candidate models. Each
model $s_i : \mathbb{R}^d \to \mathbb{R}^r$ has its own parameters (weights,
biases, etc.). The combined predictor is defined as a convex combination:
\begin{equation}\label{eq:mixture}
f(\cdot) \;=\; \sum_{i \in \mathcal{M}} \lambda_i\, s_i(\cdot),
\quad
\lambda_i \geq 0,\quad \sum_{i \in \mathcal{M}} \lambda_i = 1.
\end{equation}
This ensemble structure has two key properties, stated as a theorem.

\begin{theorem}[Tractability of the convex ensemble]\label{thm:ensemble}
Let $f = \sum_{i\in \mathcal{M}} \lambda_i s_i$ with $\lambda_i \geq 0$, $\sum_i \lambda_i = 1$.
\begin{enumerate}[(i)]
\item \emph{(Loss bound.)} If $L(\cdot,\hat{y})$ is convex in its
  first argument, then by Jensen's inequality \cite{jensen1906fonctions}:
  \begin{equation}\label{eq:jensen}
  L\!\Bigl(\sum_{i\in \mathcal{M}} \lambda_i\, s_i(\bar x),\, \hat{y}\Bigr)
  \;\leq\;
  \sum_{i\in \mathcal{M}} \lambda_i\, L\bigl(s_i(\bar x), \hat{y}\bigr).
  \end{equation}
  Minimizing the right-hand side, linear in $\lambda$ for fixed $s_i$ also
  reduces the actual mixture loss.
\item \emph{(Constraint linearization.)} For any $a\in\mathbb{R}^{r\times m}$, and fixed models $s_i$:
  \begin{equation}\label{eq:linear-ct}
  \sum_{q = 1}^ra^q_j\!\Bigl(\sum_{i\in \mathcal{M}} \lambda_i\, s_i(\bar{x}_j)\Bigr) \;\geq\; b_j
  \;\iff\;
  \sum_{i\in \mathcal{M}} \lambda_i\,\bigl(\sum_{q = 1}^ra^q_j s_i(\bar{x}_j)\bigr) \;\geq\; b_j.
  \end{equation}
  This equivalence follows from the linearity of matrix-vector
  multiplication: $\sum_{q = 1}^ra^q_j (\sum_i \lambda_i v_i) = \sum_i \lambda_i (\sum_{q = 1}^ra^q_j v_i)$ for any
  vectors $v_i \in \mathbb{R}^n$ and scalars $\lambda_i$. The output
  constraints thus become \emph{linear} in the weights $\lambda$ for fixed
  models, making the master problem tractable as an LP.
\end{enumerate}
\end{theorem}

In the following table, we summarize the notations used in the remaining of the paper. 
\begin{table}[htbp]
\centering
\caption{Summary of notations}
\label{tab:notations}

\begin{tabular}{c p{9cm}}
\toprule
\textbf{Notation} & \textbf{Description} \\
\midrule
$n$ & Size of the input data set \\
$m$ & Size of the hard constraints set \\
$d$ & Input size of the AI model \\
$r$ & Output size of the AI model \\
$\mathcal{M}$ & Set of all possible AI models (candidates) \\
$\mathcal{I}^{\text{Const}}$ & Set of inputs for which output constraints must be satisfied \\
$\mathcal{I}^{\text{Train}}$ & Set of training data \\
$f\in \mathbb R^r$ & Prediction given by the linear combination of AI models, i.e., $\sum_i \lambda_i s_i$\\
\bottomrule
\end{tabular}

\end{table}

~\\
The key observation is that (i) the loss upper bound and (ii) the
constraint linearization together reduce the original intractable
problem~\eqref{eq:clearn} to a linear program in the weights~$y$.

\subsection{Master Linear Program}\label{sub:master} 
Given a set of trained models $\{s_i : i \in \mathcal{M}\}$, the mixture weights
are determined by solving the following LP:
\begin{alignat}{3}
\min \quad
&\sum_{i \in \mathcal{M}} \mathcal{L}_i\, \lambda_i
  \;+\; M_s\sum_{j=1}^m  \delta_j
  \label{eq:master-obj}\\
\text{s.t.}\quad
&\sum_{i \in \mathcal{M}} \lambda_i = 1,
  \label{eq:master-simplex}
  \qquad\qquad[\gamma]\\
&\sum_{i \in \mathcal{M}} \lambda_i \sum_{q = 1}^ra^q_j s^q_i(\bar{x}_j) \;+\; \delta_j \;\geq\; b_j,
  \qquad \forall j = 1,\ldots,m,
  \label{eq:master-ct}
  \quad [\beta_j]\\
&\lambda_i \geq 0,
  \qquad\qquad \forall i \in \mathcal{M},\\
&\delta_j \geq 0,
  \qquad\qquad \forall j \in 1,\dots,m,
  \label{eq:master-nn}
\end{alignat}
where $\mathcal{L}_i = \sum_{k=1}^n L(s_i(\bar{x}_k),\hat{y}_k)$
is the training loss of model $i$ on $\mathcal{I}^{Train}$, $M_s \gg 1$
is a large feasibility penalty.
The slack variable $\delta \in \mathbb{R}_+^m$ is an
$m$-dimensional vector (one component per hard constraint); the term
$\delta$ aggregates violations into a scalar penalty.
In our applications (Sections~\ref{sec:mnist} and~\ref{sec:mcf}),
each constraint is scalar ($m=1$), so $\delta_j \in \mathbb{R}_+$.
Dual variables $\gamma \in \mathbb{R}$ (equality) and $\beta \in \mathbb{R}_+^m$ ($\geq$ inequality)
are associated with constraints~\eqref{eq:master-simplex}
and~\eqref{eq:master-ct}, respectively.

\textbf{Design of the slack variables $\delta_j$.}
When the constraint set $\mathcal{I}^{Const} = \{\bar{x}_1,\ldots,\bar{x}_m\}$
is given explicitly, we introduce one slack $\delta_j \geq 0$ per input
$\bar{x}_j$, yielding $m$ slack variables. When $\mathcal{I}^{Const}$ is large
or continuous and constraints are added incrementally by the cutting-plane
procedure (Algorithm~\ref{alg:cgcp}), we use a \emph{single} shared slack
variable representing the worst-case violation over all active constraints.
This avoids introducing a new variable for each
generated constraint and keeps the master LP compact as the cutting-plane
loop progresses. The shared slack is sufficient because the separation
step (Eq.~\eqref{eq:separation}) identifies the single most violated input, so
only one slack needs to be driven to zero at each outer iteration.

Constraint~\eqref{eq:master-simplex} requires that the mixture weights
form a valid convex combination; its dual variable $\gamma$ represents
the marginal value of adding one more model to the ensemble and appears
in the reduced cost. Constraint~\eqref{eq:master-ct}
requires that the ensemble's weighted prediction satisfies the output
constraints at each point $\bar{x}_j \in \mathcal{I}^{Const}$; its dual variable
$\beta_j \in \mathbb{R}_+^m$ quantifies how difficult it is to satisfy constraint $j$
given the current columns, a large $\beta_j$ signals a highly violated
constraint and increases the weight placed on input $\bar{x}_j$ during
the next pricing step.

When all slacks are zero, the ensemble satisfies all constraints in
$\mathcal{I}^{Const}$. The LP is always feasible (slacks absorb any violation),
and the penalty $M_s$ ensures that feasibility is prioritized over loss
minimization.

\begin{remark}[Single-model selection via branching]
If one needs to enforce constraints with a \emph{single} model (e.g.\
for interpretability or deployment simplicity), it suffices to replace
the continuous weights $\lambda_i \in [0,1]$ with binary variables
$\lambda_i \in \{0,1\}$. This forces the LP to select exactly one column,
turning CG4AI into a constrained model-selection problem solvable by
branch-and-bound on top of the LP relaxation.  The branching strategy must be carefully designed due to the column generation algorithm.  This provides a Branch-and-price algorithm. The LP relaxation studied here already provides strong
constraint guarantees and is tractable; branch-and-bound is deferred
to future work.
\end{remark}

\subsection{Column Generation}\label{sub:cg}

The set $\mathcal{M}$ is typically exponentially large. Column generation~\citep{barnhart1998branch,demiriz2002lpboost} maintains
a restricted master LP (RMP) with a subset $\mathcal{M}' \subseteq \mathcal{M}$ of active
models and iteratively adds columns (models) with negative reduced cost.
The \emph{reduced cost} of model $i$ is:
\begin{equation}\label{eq:rc}
r_i \;=\; \mathcal{L}_i \;-\; \sum_{j=1}^m \beta_j^\top \, \sum_{q = 1}^ra^q_j s_i^q(\bar{x}_j) \;-\; \gamma,
\end{equation}
where $\beta_j$ and $\gamma$ are the dual variables of the current RMP
solution. A model with $r_i < 0$ can improve the objective and should
be added to the ensemble.\\

The \emph{pricing problem} finds the model that minimizes the reduced
cost:
\begin{equation}\label{eq:pricing}
\min_{w \in W}\;
r(w) \;=\;
\mathcal{L}(w)
\;-\;\sum_{j=1}^m \beta_j^\top \sum_{q = 1}^ra^q_j s_w^q(\bar x_j)
\;-\;\gamma,
\end{equation}
where $w$ denotes the model weights. This is a training problem: we
seek a model with low training loss that also helps satisfy the
constraints. The dual variable $\beta_j$ controls how much weight is
placed on constraint $j$: inputs with large $\beta_j$ receive more
attention during pricing, focusing training on the most violated
constraints.

Problem~\eqref{eq:pricing} is dependent on the considered AI model. In practice, libraries like PyTorch and scikit-learn provide efficient algorithms for the learning of multiple types of AI models. When $s_w$ is differentiable, the reduced cost is
differentiable with respect to $w$ and standard backpropagation applies.
For linear (affine) models, problem~\eqref{eq:pricing} is convex in
$w$ and the optimizer finds a global minimum, so the CG stopping
criterion is theoretically valid. For nonlinear models (e.g., networks
with ReLU activations), the pricing is non-convex and a local optimizer
finds only a local minimum; in this case Algorithm~\ref{alg:cg} is a
\emph{heuristic} that generates improving columns as long as the local
search finds one, without the LP optimality guarantee of exact CG.
The full algorithm is given in Algorithm~\ref{alg:cg}.

\begin{remark}[Convergence guarantees]\label{rem:conv}
Three convergence regimes arise depending on how the pricing is solved.
(1)~\emph{Master Linear Program (MLP) columns with MIP pricing.} The pricing can be reformulated
as a MIP
\citep{aftabi2024ffnnmip,anderson2020strong};
a global minimizer is again available, restoring LP optimality guarantees.
(2)~\emph{MLP columns with local optimization.} The most practical case.
No optimality certificate is available; the algorithm is a heuristic that
stops when the local search fails to reduce the reduced cost below
$-\varepsilon$. The optimality gap can be bounded a posteriori by tracking
the best reduced cost found across multiple restarts, or by solving the
pricing MIP on the final iteration. Experiments in Section~\ref{sec:mcf}
show that regime~(2) achieves feasibility and good accuracy on 38
SNDLIB instances, suggesting that local minima of the pricing are not
a major obstacle in practice.
\end{remark} 

\begin{algorithm}[ht]
\caption{CG4AI: Column Generation for Constrained Ensembles}
\label{alg:cg}
\begin{algorithmic}[1]
\State Initialize $\mathcal{M}' \leftarrow \emptyset$. Add a feasibility dummy
       column to $\mathcal{M}'$ (large loss, zero constraint contribution).
\Repeat
  \State \textbf{Master step.} Solve the RMP over $\mathcal{M}'$ to obtain dual
         variables $\beta_j$, $\gamma$ and primal weights $\lambda_i$.
  \State \textbf{Pricing step.} Train a new model $s_{\mathrm{new}}$
         by minimizing~\eqref{eq:pricing}, guided
         by $\beta_j$.
  \If{$r(w_{\mathrm{new}}) < -\varepsilon$}
    \State Add $s_{\mathrm{new}}$ to $\mathcal{M}'$.
  \EndIf
\Until{$r(w_{\mathrm{new}}) \geq -\varepsilon$ \textbf{and}
       $\sum_j \delta_j = 0$ (all constraints satisfied)}
\State \textbf{Output.} Ensemble $f = \sum_{i \in \mathcal{M}'} \lambda_i\, s_i$ with
       weights from the final RMP.
\end{algorithmic}
\end{algorithm}

The algorithm stops when no column with negative reduced cost exists
\emph{and} the current solution is feasible. If the column limit is
reached before feasibility, the output is still a well-defined convex
combination, but some constraints may not be satisfied; the total
violation is measured by $\sum_j \delta_j$.

\subsection{Cutting Planes for Large Constraint Sets}\label{sub:cutting}

When $\mathcal{I}^{Const}$ is large or continuous, it is impractical to include
all constraints in the RMP. We use a cutting-plane
approach~\citep{kelley1960cutting}: start with a small subset
$\mathcal{I}^{Const}_0 \subseteq \mathcal{I}^{Const}$, and iteratively add the
constraint corresponding to the most violated input.

Given the current ensemble weights $\lambda^*$, the \emph{separation problem}
is:
\begin{equation}\label{eq:separation}
\bar{x}^* \;=\;
\arg\max_{\bar{x} \in \mathcal{I}^{Const}}\;
\max_{j = 1,\ldots,m}
\Bigl(c_j - \sum_{i \in \mathcal{M}'} \lambda_i^*\, \sum_{q = 1}^ra^q_j s_i^q(\bar{x})\Bigr),
\end{equation}
This finds the input in
$\mathcal{I}^{Const}$ that is most violated under the current ensemble. If the
maximum is non-positive, the ensemble is feasible for all of
$\mathcal{I}^{Const}$. In general the separation problem is nonconvex and is
solved heuristically by a black-box optimizer. Algorithm~\ref{alg:cgcp}
combines column generation with cutting planes.\\

\begin{algorithm}[ht]
\caption{CG4AI with Cutting Planes}
\label{alg:cgcp}
\begin{algorithmic}[1]
\State Initialize $\mathcal{I}^{Const}_{\mathrm{RMP}} \leftarrow \mathcal{I}^{Const}_0$
       (small initial constraint set).
\Repeat
  \State Run Algorithm~\ref{alg:cg} with $\mathcal{I}^{Const}_{\mathrm{RMP}}$
         to get mixture weights $\lambda^*$.
  \State \textbf{Separation step.} Solve~\eqref{eq:separation} to find
         the most violated input $\bar{x}^*$.
  \If{$\Bigl(c_j- \sum_{i \in \mathcal{M}'} \lambda_i^*\, \sum_{q = 1}^ra^q_j s_i^q(\bar{x}^*)\Bigr))
 > \varepsilon$}
    \State Add $\bar{x}^*$ to $\mathcal{I}^{Const}_{\mathrm{RMP}}$.
  \EndIf
\Until{no violated input found}
\State \textbf{Output.} Ensemble $f = \sum_{i \in \mathcal{M}'} \lambda^*_i\, s_i$.
\end{algorithmic}
\end{algorithm}

The combined procedure alternates between pricing (new columns), solving
the master LP (new weights), and separation (new constraints). When all
three steps report no improvement, the solution is both LP-optimal and
feasible for the full set $\mathcal{I}^{Const}$.

\paragraph{MIP-based exact separation.}
When the columns $s_i$ are single-hidden-layer ReLU networks, both the
pricing and the separation problems can be solved to global optimality
via mixed-integer programming~\citep{aftabi2024ffnnmip,anderson2020strong}.
For separation specifically, the MIP encodes the ensemble output
constraints and searches for the input $\bar{x}^* \in \mathcal{I}^{Const}$ that
maximally violates them. This yields
\emph{provably optimal cuts}, tightening the master LP as much as
possible at each outer iteration. In the MCF application
(Section~\ref{sec:mcf}), the black-box NLopt optimizer \cite{johnson2014nlopt} suffices
empirically; the MIP approach provides a path to certified optimality
for applications requiring formal guarantees on the cutting-plane
procedure.

\section{Proof of Concept: Digit Classification}\label{sec:mnist}

We illustrate the CG4AI mechanisms on MNIST~\citep{lecun1998gradient}
using a deliberately \emph{minimalist} MLP ($h=4$ hidden neurons).
This choice is intentional: with only 4 neurons, a single model achieves
modest accuracy ($52{-}77\%$ depending on training size), making the
accuracy gains from adding constraints clearly measurable. It also
represents the regime of embedded or resource-constrained systems where
model capacity is limited by design. More expressive architectures would
reduce the absolute gains but the constraint-enforcement mechanism is
architecture-agnostic. The section is didactic and concise; the main
quantitative results are in Section~\ref{sec:mcf}.

\subsection{Column Model and Constraint Formulations}\label{sub:mnist-setup}

Let us introduce $\lambda_i s_i^c(x)$ the probability of the class $c$ associated to input $x$ on the model $i$.  For an input $x_j$, $c_j$ is the right class of the input $x_j$.    

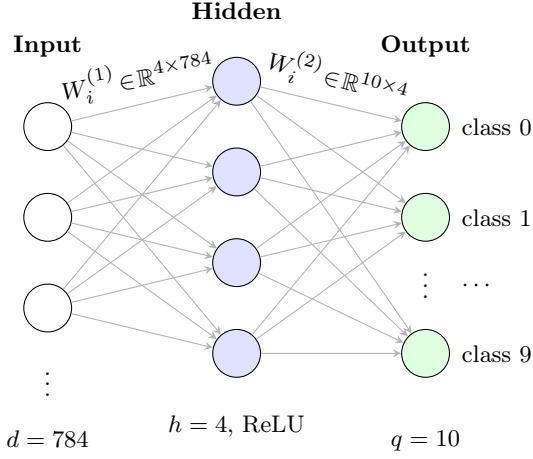
\begin{figure}[ht]
\centering
\begin{tikzpicture}[
    node distance = 0pt,
    neuron/.style={circle, draw, minimum size=18pt, inner sep=0pt,
                   fill=white, font=\small},
    dots/.style={font=\small},
    layer/.style={font=\small\bfseries},
    >=stealth
]
\foreach \i/\y in {1/1.2, 2/0, 3/-1.2}{
    \node[neuron] (M\i) at (0, \y) {};
}
\node[dots] at (0,-2.1) {$\vdots$};
\node[layer, above] at (0, 2.0) {Input};
\node[font=\scriptsize, below] at (0,-2.7) {$d=784$};

\foreach \i/\y in {1/1.8, 2/0.6, 3/-0.6, 4/-1.8}{
    \node[neuron, fill=blue!12] (H\i) at (2.5, \y) {};
}
\node[layer, above] at (2.5, 2.5) {Hidden};
\node[font=\scriptsize, below] at (2.5,-2.5) {$h=4$, ReLU};

\foreach \i/\y in {1/1.2, 2/0, 3/-1.8}{
    \node[neuron, fill=green!12] (O\i) at (5, \y) {};
}
\node[dots] at (5,-0.8) {$\vdots$};
\node[layer, above] at (5, 2.0) {Output};
\node[font=\scriptsize, below] at (5,-2.7) {$q=10$};

\foreach \i in {1,2,3}{
    \foreach \j in {1,2,3,4}{
        \draw[->, gray!60, thin] (M\i) -- (H\j);
    }
}
\foreach \i in {1,2,3,4}{
    \foreach \j in {1,2,3}{
        \draw[->, gray!60, thin] (H\i) -- (O\j);
    }
}
\foreach \i/\c in {1/0, 2/1, 3/9}{
    \node[font=\scriptsize, right=4pt] at (O\i) {~~class \c};
}
\node[font=\scriptsize, right=4pt] at (5,-0.9) {$~~\cdots$};

\node[font=\scriptsize, above, rotate=10] at (1.25, 1.5)
    {$W_i^{(1)}\!\in\!\mathbb{R}^{4\times784}$};
\node[font=\scriptsize, above, rotate=-15] at (3.75, 1.5)
    {$W_i^{(2)}\!\in\!\mathbb{R}^{10\times4}$};
\end{tikzpicture}
\caption{Architecture of column model $s_i$ used in the MNIST
         experiments: one hidden layer with $h=4$ neurons and ReLU
         activation, mapping a $784$-dimensional input to $10$ class
         logits. The ensemble output is $\sum_{i\in \mathcal M} \lambda_i s_i(x)$.}
\label{fig:mlp-tikz}
\end{figure}

\paragraph{Architecture}
Each column maps $x\!\in\!\mathbb{R}^{784}$ to 10 logits via
$s_i(x) = W_i^{(2)}\mathrm{ReLU}(W_i^{(1)}x+b_i^{(1)})+b_i^{(2)}$,
with $W_i^{(1)}\!\in\!\mathbb{R}^{4\times784}$, $W_i^{(2)}\!\in\!\mathbb{R}^{10\times4}$.
The ensemble predicts $\arg\max_c\sum_i \lambda_i s_i^c(x)$.
Columns are trained with Adam ($\eta=10^{-3}$, up to 1000 epochs,
early stopping at patience 50). Figure~\ref{fig:mlp-tikz} shows the architecture.

When the constraint matrix $A$ encodes per-sample classification requirements,
the feasibility of the LP translates into guarantees on~$\mathcal{I}^{Const}$.
The two formulations below each use their constraint coefficient as the
ensemble prediction signal, making the guarantee self-consistent and the
proof immediate.\\
For a point $j\in \{1,...,m\}$:

\begin{itemize}
    \item \textbf{CG-proba}: require $\sum_{i\in \mathcal M} \lambda_i\,(s_i^{c_j}(\bar x_j))\geq\delta$,  it corresponds to an absolute probability needs. Thus if $\delta>0.5$ then $\bar x_j$ is classified in $c_j$. 
    \item \textbf{CG-margin}: require $\sum_{i\in \mathcal M} \lambda_i[(s_i^{c_j}(\bar x_j))-\max_{c\neq c_j}s_{i}^{c}(\bar x_{j})]\geq\delta$,  it corresponds to a relative probability needs. Thus if $\delta>0$ then the probability to classify $\bar x_j$ in $c_j$ is better than all other classes. 
\end{itemize}

\subsection{Scenarios}\label{sub:mnist-scenarios}

\paragraph{Scenario 1 — Hard set / optimization set (key feature).}

The LP enforces hard constraints on inputs in hard set $\mathcal{I}^{Const}$ while the pricing minimizes cross-entropy on optimization set $\mathcal{I}^{Train}$. When $\mathcal{I}^{Train}=\emptyset$
the model learns entirely from the dual values. We test four splits of the
per-class budget: $(0,10), (2,8), (5,5), (8,2)$ where the first number is the size of the Hard Set and the second is the size of the Optimization Set. 
This scenario allows considering a hierarchy between two sets of data.

\begin{figure}[ht]
    \centering
    \includegraphics[width=0.198\linewidth]{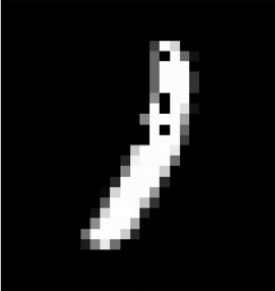}
    \includegraphics[width=0.195\linewidth]{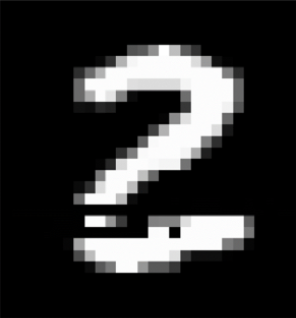}
    \includegraphics[width=0.21\linewidth]{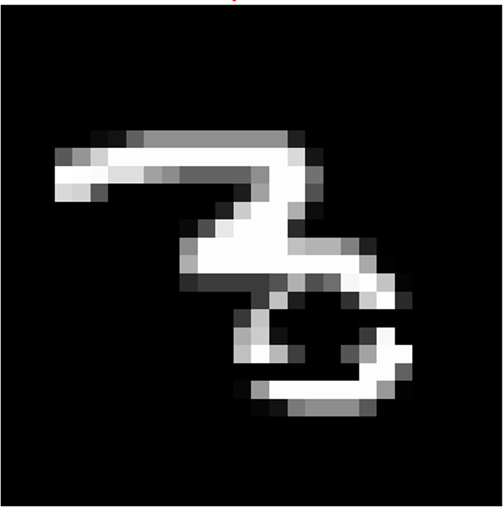}
    \caption{Three adversarially perturbed digits (up to 100 pixels
             flipped). The  models before adding constraints predict 2, 3, and 4
             respectively, but the original labels are different.
             CG4AI adds constraints to enforce correct classification
             of these perturbed inputs.}
    \label{fig:modified_digit}
\end{figure}

\paragraph{Scenario 2 — Adversarial robustness (CG-robust).}
After training a base column, greedy saliency-based pixel flips (up to $P$ pixels;
$P=100$ in our experiments, but any black-box attack can replace this)
generate adversarial perturbations $\tilde x_j$.
The salience score is the modification of probability when a pixel changes.  
Each misclassified $\tilde x_j$ adds the constraint CG-proba or CG-margin 
to $\mathcal{I}^{Const}$. A black-box optimizer searching for the minimum perturbation
that breaks the prediction can tighten these constraints and reduce
cutting-plane iterations.
Figure~\ref{fig:modified_digit} shows example perturbed digits. Cutting plane is used to generate constraints.

This scenario allows us to check if there exists an AI model able to classify digits and be robust to the greedy saliency-based pixel flips. Note that this approach can be considered only with optimization approach $\mathcal{I}^{Train}$. 

\begin{figure}[ht]
    \centering
    \includegraphics[width=0.2\linewidth]{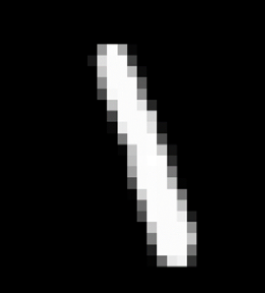}
    \includegraphics[width=0.2\linewidth]{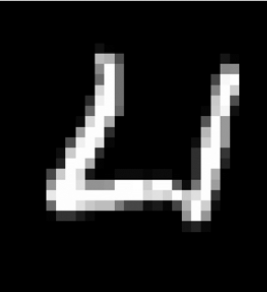}
    \caption{Two training images initially misclassified: digit~1
             is predicted as~3, and digit~4 is predicted as~0.
             Adding hard constraints for these images forces the
             ensemble to classify them correctly.}
    \label{fig:wrong_prediction}
\end{figure}

\paragraph{Scenario 3 — Correcting misclassified examples (CG-mis).}
Let $\mathcal{E}$ be training images misclassified by the base column.
Adding  the constraint CG-proba or CG-margin
for each element of $
\mathcal{E}$ forces the ensemble to correct them.
Figure~\ref{fig:wrong_prediction} shows two corrected digits.

\begin{figure}[ht]
    \centering
    \includegraphics[width=0.195\linewidth]{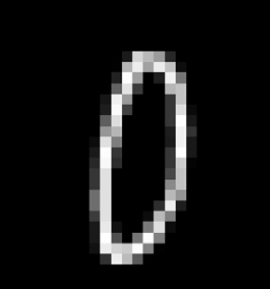}
    \includegraphics[width=0.2\linewidth]{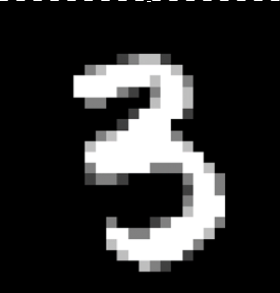}
    \includegraphics[width=0.205\linewidth]{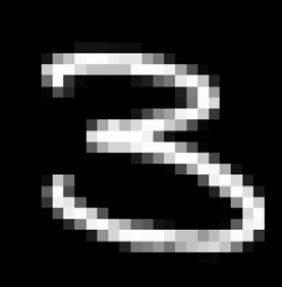}
    \includegraphics[width=0.205\linewidth]{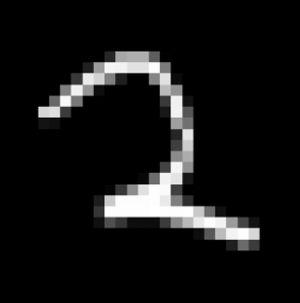}
    \caption{Relabeling example: three digits (0, 3, 3, 2) are
             shown after adding constraints forcing digit~3 to be
             predicted as~2. The digit-0 image is predicted as 0 by
             50\%, as 2 by 42\%, and as 4 by 8\%. The first 3 is
             predicted as 2 by 54\%, the second as 2 by 100\%, and
             the 2 as 2 by 100\%.}
    \label{fig:force_output}
\end{figure}

This scenario guarantees that misclassified input spotted by a user for instance must be well classified.
\paragraph{Scenario 4 — Output relabeling (CG-relabel).}
To force class $c'$ for all inputs in a target set $I'\subseteq I$, add
$\sum_{i\in \mathcal M} \lambda_i\,s_i^{c'}(\bar x_j)\geq\theta$ 
for each $j\in I'$. The original training labels are unchanged;
the LP imposes the relabeling through new constraint rows, usually
requiring only a few additional columns.
Figure~\ref{fig:force_output} shows digit-3 images relabeled as digit 2.

This last scenario guarantees some inputs are relabeled, it is interesting, for instance, when some classes must be merged. 

\subsection{Results}\label{sub:mnist-results}

Training sizes range from $n=10$ to $n=1000$ (balanced across 10 classes).
Test set: 2000 images (200 per class). LP solver: CPLEX 12.6.3 \cite{cplex1263}; column limit: 400.

\begin{figure}[ht]
\centering
\includegraphics[width=0.7\linewidth]{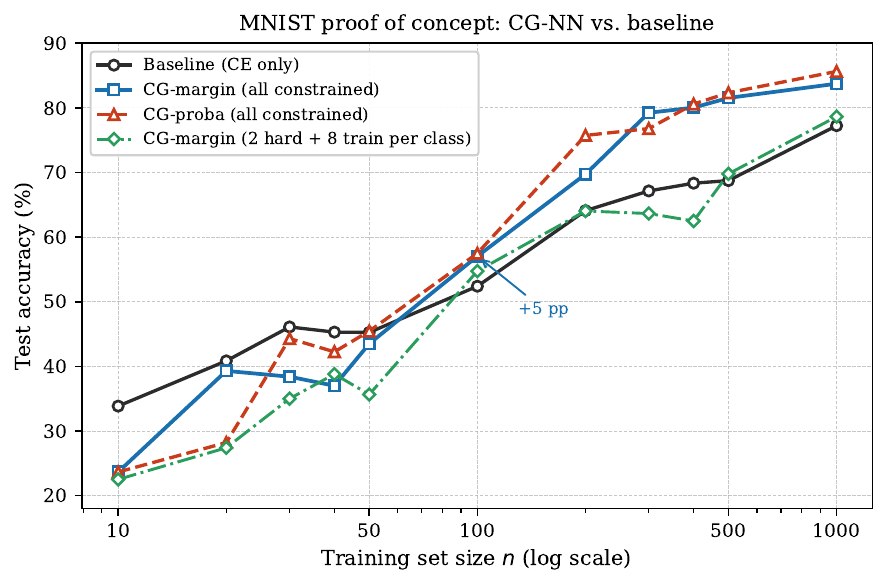}
\caption{Test accuracy on 2000 MNIST images vs.\ training set size.
         CG-margin and CG-proba (constraints only, $k_o=0$) outperform
         the baseline at every scale. The ``2 hard + 8 train'' split
         provides a guaranteed-correct hard set with a small accuracy cost.}
\label{fig:mnist-summary}
\end{figure}

\begin{figure}[ht]
\centering
\includegraphics[width=\linewidth]{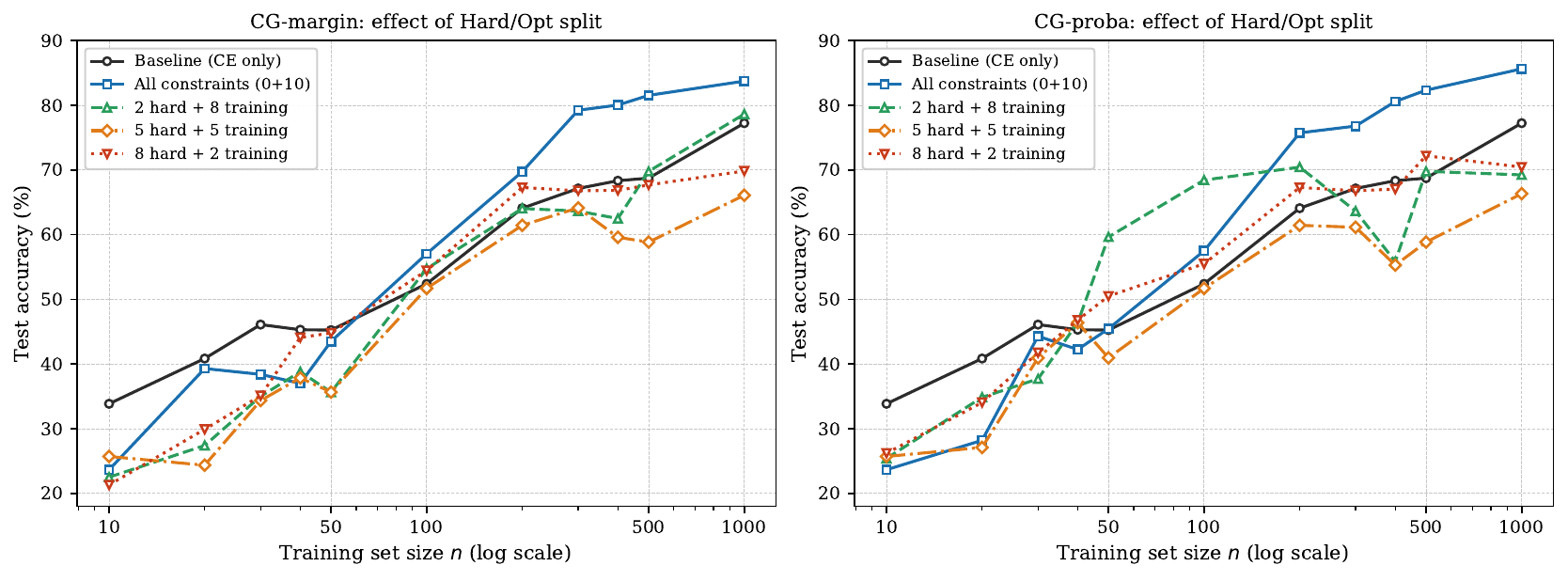}
\caption{Effect of the hard/optimization split on test accuracy
         (CG-margin left, CG-proba right). All CG4AI variants surpass
         the baseline. Increasing $k_h$ extends the guarantee coverage
         but reduces overall test accuracy.}
\label{fig:mnist-hardopt}
\end{figure}

\begin{table}[ht]
\centering
\caption{Test accuracy (\%) on 2000 MNIST images (MLP, $h=4$, all-constrained
         mode unless stated). $k_h$: images per class in hard set $\mathcal{I}^{Const}$.
         Single run (seed 42). Note: the comparison in Table~\ref{tab:comp-acc}
         uses a separate 3-seed benchmark (seeds 42, 43, 44), giving a slightly
         different baseline ($56.1\%$ at $n=100$) due to seed variability.}
\label{tab:mnist-acc}
\small
\begin{tabular}{llcc}
\toprule
Variant & Split & $n=100$ & $n=1000$ \\
\midrule
Baseline (CE only)    & —             & 52.4 & 77.3 \\
\midrule
CG-margin             & all constrained & 57.1 & 83.8 \\
CG-proba              & all constrained & 57.5 & 85.7 \\
CG-margin             & $k_h=2$, rest optimize & 54.8 & 78.7 \\
CG-proba              & $k_h=2$, rest optimize & 68.5 & 69.3 \\
CG-margin             & $k_h=5$, rest optimize & 51.7 & 66.1 \\
\midrule
CG-robust             & —             & 60.0 & 76.7 \\
CG-mis                & —             & 61.3 & 75.9 \\
CG-relabel$^\dagger$  & —             & 38.9 & 69.3 \\
\bottomrule
\multicolumn{4}{l}{\small $^\dagger$ Evaluated on original labels (digit~3$\to$2 relabeling).}
\end{tabular}
\end{table}

All CG4AI all-constrained variants outperform the baseline at every training size.
At $n=1000$, CG-proba reaches $85.7\%$ ($+8.4$~pp) of accuracy. With $k_h=2$, CG-margin
reaches $78.7\%$ (above baseline) while the LP guarantees that every image in the
hard set $\mathcal{I}^{Const}$ is correctly classified by the ensemble whenever the
corresponding LP constraint is satisfied a formal classification guarantee
no penalty-based or projection method can provide. CG-robust and CG-mis surpass
the baseline despite allocating column budget to adversarial and correction constraints.

\subsection{Comparison with Constrained-Learning Baselines}\label{sub:mnist-comparison}

Competing methods (3-seed averages): \textbf{Penalty (L2)}: CE~$+~10\|\text{viol}\|^2$; \textbf{Aug.\ Lagrangian}: dual-ascent on per-image multipliers;
\textbf{DC3}~\citep{donti2021dc3}: $K=5$ unrolled gradient-correction steps;
\textbf{Projection}: inference-time L2-projection using true test labels
(oracle — shown for reference only). Same MLP for all.

\begin{table}[ht]
\centering
\caption{Test accuracy (\%) at $n=100$ and $n=1000$, averaged over 3
         seeds (baselines) and single C++ run (CG4AI). Same MLP for
         all methods (1 hidden layer, 4 neurons). Best result excluding
         the oracle Projection is \textbf{bolded}.}
\label{tab:comp-acc}
\small
\begin{tabular}{lcc}
\toprule
Method & $n=100$ & $n=1000$ \\
\midrule
Baseline (single model, CE)  & 56.08 & 70.98 \\
Penalty (L2), $\lambda=10$   & 16.18 & 52.00 \\
Augmented Lagrangian         & 18.80 & 61.35 \\
DC3 ($K=5$ steps)            & 56.70 & 71.73 \\
Projection$^\dagger$ (oracle)& 82.37 & 90.42 \\
\midrule
\textbf{CG-margin}           & \textbf{69.70} & \textbf{86.30} \\
\textbf{CG-proba}            & \textbf{68.55} & \textbf{86.05} \\
\bottomrule
\end{tabular}
\end{table}

\begin{table}[ht]
\centering
\caption{Training constraint satisfaction (\%) at $n=100$ and
         $n=1000$, 3-seed averages for baselines. CG-proba provides
         a formal LP guarantee; all others are heuristic.}
\label{tab:comp-sat}
\small
\begin{tabular}{lcc}
\toprule
Method & $n=100$ & $n=1000$ \\
\midrule
Baseline              &  99.7 &  99.7 \\
Penalty (L2)          &   7.0 &  11.2 \\
Augmented Lagrangian  &   9.3 &  16.7 \\
DC3                   &  99.7 &  99.7 \\
\midrule
\textbf{CG-margin}  & \textbf{100.0} & 69.7$^\dagger$ \\
\textbf{CG-proba}   & \textbf{100.0} & \textbf{100.0} \\
\bottomrule
\multicolumn{3}{l}{\small $^\ast$ LP constraint satisfied $\Rightarrow$ correct classification (each under its prediction rule).}\\
\multicolumn{3}{l}{\small $^\dagger$ 400-column limit; 69.7\% of constraints satisfied.}\\

\end{tabular}
\end{table}

\begin{figure}[ht]
    \centering
    \includegraphics[width=\linewidth]{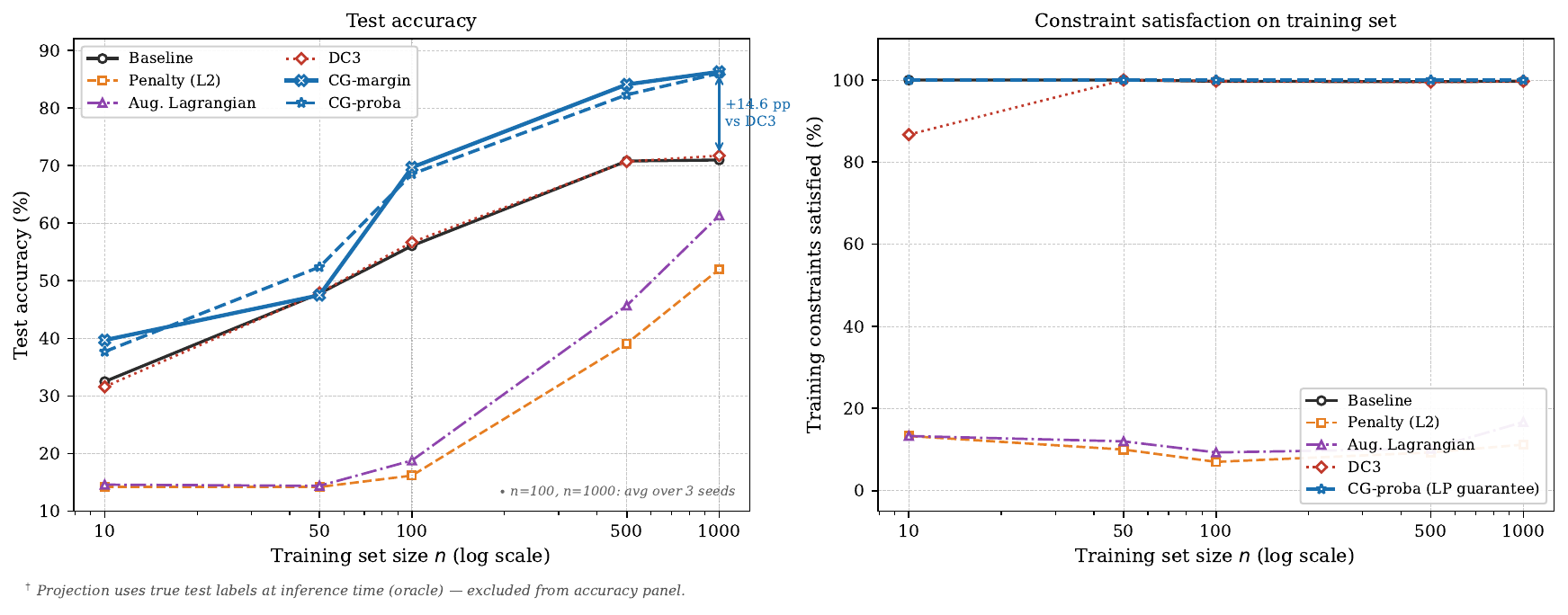}
    \caption{Test accuracy (left) and training constraint satisfaction
             (right). DC3 and the baseline are nearly superimposed,
             confirming that constraint-correction without an LP guarantee
             provides negligible accuracy gain. Both CG4AI variants
             separate clearly from all competitors; the $+14.6$~pp gap
             over DC3 at $n=1000$ is the measurable effect of the LP
             output constraints. Penalty and Aug.\ Lagrangian satisfy
             fewer than $17\%$ of constraints at any scale.}
    \label{fig:bench-comparison}
\end{figure}

DC3 gains only $+0.62$~pp ($n=100$) and $+0.75$~pp ($n=1000$) over the baseline, 
our proxy for ensembling without LP output constraints (cf.\ LPBoost).
CG-margin exceeds DC3 by $+13.0$ and $+14.6$~pp: this gap is the direct
effect of the LP constraint rows~\eqref{eq:master-ct} and their dual-guided pricing.
Penalty methods collapse to $\leq 18.8\%$ at $n=100$ (below the baseline of $56.1\%$),
confirming that soft constraints are unreliable with limited model capacity.
Both formulations provide a formal classification guarantee when the LP
constraint is satisfied,
each under its consistent prediction rule. CG-proba reaches full LP feasibility
at every training size (100\% of constraints, averaged-probability prediction).
CG-margin achieves 69.7\% constraint satisfaction at $n=1000$ (400-column limit
reached); the remaining 30.3\% are not formally guaranteed under the averaged-logit
prediction rule. CG-proba is recommended when complete coverage is required;
CG-margin when maximum test accuracy is the priority.
~~\\\\
~~\\\\
~~\\\\

\newpage

\section{Main Use Case: Network Routing Under Capacity Constraints}\label{sec:mcf}

Section~\ref{sec:mnist} established CG4AI's mechanisms on a simple
classification task: a constraint forces correct classification of a
designated input, the LP dual measures how much the constraint is
violated, and the pricing trains the next column to address that
violation. The same mechanism now operates on a realistic combinatorial
optimization problem where constraints arise not from user designations
but from physics: link capacity limits in a telecommunication network.

The key difference from Section~\ref{sec:mnist} is scale and structure.
The constraint set $\mathcal{I}^{Const}$ is continuous (all possible demand
vectors), so the cutting-plane procedure (Algorithm~\ref{alg:cgcp}) is
essential: at each outer iteration, a black-box optimizer finds the
demand vector $b^*$ that maximally leading to violations in
capacity constraints, adds it to $\mathcal{I}^{Const}$, and triggers another
column-generation round. The dual variables $\beta_b^a$ quantify how
much arc $a$ is overloaded under a demand scenario, directing each new
network column toward routing decisions that respect capacity.

\subsection{Problem Definition}\label{sub:mcf-problem}

We apply CG4AI to the Multi-Commodity Flow Problem (MCFP), a natural
testbed for hard constraints because the feasibility requirements
(capacity limits) are well-defined and computationally non-trivial to
satisfy with a learned model.

We model the communication network as a directed graph $G = (V, A)$,
where $V$ is the set of nodes (routers) and $A$ is the set of arcs
(links). Each link $(ij)\in A$ has a limited bandwidth capacity $c_{ij}\in \mathbb R^+$. We are given a set of demands $D$. Each demand $d \in D$ is
characterized by a source, a destination, and a traffic
volume $b^d$. Let $\mathcal{P}_d$ be a set of paths connecting the source of to the destination of $d$. The goal is to route all demands while minimizing the
\emph{maximum link utilization} (MLU):
\begin{alignat}{3}
\label{eq:mcfp1}\min\quad & MLU \\
\label{eq:mcfp2}\text{s.t.} \quad
& \sum_{p \in \mathcal{P}_d} f_p^d = 1, && & \forall d \in D, \\
\label{eq:mcfp3}& \sum_{d \in D}\; \sum_{\substack{p \in \mathcal{P}_d \\ (i,j) \in p}}
  b^d\, f_p^d \;\leq\; c_{ij}\,MLU, && & \forall (i,j) \in A, \\
\label{eq:mcfp4}& f_p^d \geq 0, && & \forall p \in \mathcal{P}_d,\;\forall d \in D,
\end{alignat}
where $MLU$ is the maximum link utilization and $f_p^d$ is the fraction of demand $d$
routed on path $p\in \mathcal{P}_d$. The first constraint ensures that each demand is
fully routed, the second enforces capacity relative to $MLU$, and
the third ensures non-negativity. Minimizing $MLU$ is equivalent to
minimizing the maximum link utilization.

\subsection{Learning Routing Splits for MCFP} \label{section_learning_mcf}

Instead of solving~\eqref{eq:mcfp1}-\eqref{eq:mcfp4} for each new demand vector, one may train
AI models to predict the routing splits $f_p^d$ directly. For a given
demand vector $b = (b^d)_{d \in D}$, the model outputs predicted splits
$\hat{f}_p^d \geq 0$ with $\sum_p \hat{f}_p^d = 1$ for each demand $d$
(the demand-satisfaction constraint is enforced by a softmax output
layer). The model is trained on a dataset of $(b, f^\star)$ pairs, where
$f^\star$ is the optimal routing obtained by solving~\eqref{eq:mcfp1}-\eqref{eq:mcfp4}.
The loss function is:
\begin{equation}
\mathcal{L}(w) = \sum_{d \in D}\sum_{p \in \mathcal{P}_d}
\bigl(\hat{f}_p^d(b; w) - f_p^{\star d}(b)\bigr)^2.
\end{equation}
Once trained, the model can produce routing splits for new demand
vectors in milliseconds, without solving an LP. However, pure neural
network predictions may violate capacity constraints, especially for
high-utilization scenarios (MLU close to 1).

Figure~\ref{fig:small_violations} illustrates this issue: for instances
where $0.99 < \text{MLU} < 1$, the neural network often produces splits
that exceed capacity.

\begin{minipage}{1\textwidth}
    \centering
    \includegraphics[width=0.8\linewidth]{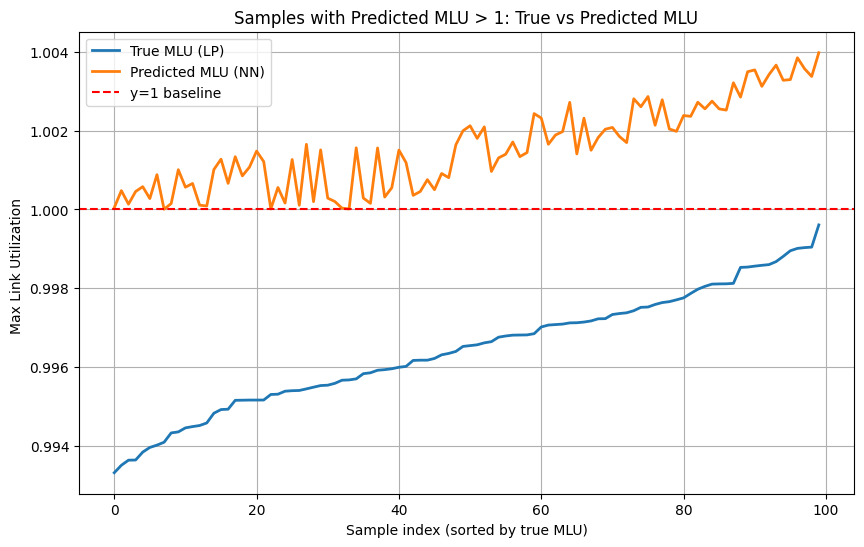}
    \captionof{figure}{Constraint violations near the MLU boundary: for
    high-utilization scenarios ($0.99 < \text{MLU} < 1$), a plain neural
    network predictor regularly exceeds the link capacity limit.}
    \label{fig:small_violations}
\end{minipage}
~\\

\subsection{CG4AI Formulation for MCFP}
In contrast with Section \ref{section_learning_mcf}, in the following we use the CG4AI framework, introduced in Section~\ref{sec:framework}, for MCF problem. The columns $s_i$ is an AI model predicting routing splits. 
The cutting-plane procedure (Algorithm~\ref{alg:cgcp}) extends the guarantee
to continuous demand sets. For $d\in D$, let $\mathcal{M}^d$ be the set of prediction models associated to demand $d$. Each model of $\mathcal{M}^d$ has $|D|$ inputs corresponding to the bandwidth demands and $\mathcal P^d$ corresponding to the split on each path of the demand $d$.\\  

The master problem for the MCFP is as follows:
\begin{align}
\min \quad &
    \sum_{d \in D}\sum_{i \in \mathcal{M}^d} \mathcal{L}_i^d\, \lambda_i^d
    \label{eq:mix-obj}\\
\text{s.t.} \quad
& \sum_{i \in \mathcal{M}^d} \lambda_i^d = 1,
    && \forall d \in D, 
    \label{eq:mix-simplex}\\
& -\sum_{d \in D} b^d \sum_{i \in \mathcal{M}^d} \lambda_i^d\sum_{\substack{p \in \mathcal{P}^d \\ a \in p}}
   s_{i}^p(b)\,
  \;\geq\; -c_a,
    && \forall a \in A,\;\forall b \in \mathcal{I}^{Const},
    \label{eq:mix-cap}\\
& \lambda_i^d \geq 0,
    && \forall d \in D,\;\forall i \in \mathcal{M}^d.
    \label{eq:mix-nn}
\end{align}
Constraint~\eqref{eq:mix-simplex} enforces a valid convex combination of models in $\mathcal{M}^d$ per demand.
Constraint~\eqref{eq:mix-cap} enforces link capacity for all demand
vectors $b \in \mathcal{I}^{Const}$. \\
One slack variable per demand is introduced at equation \eqref{eq:mix-simplex} penalized in the objective function. It allows to keep the RMP feasible during the column generation algorithm. 
\paragraph{Pricing.}
For demand $d$ and model $i\in \mathcal{M}^d$, the reduced cost is:
\begin{equation}\label{eq:rc-mcf}
r_i^d(w_i^d) \;=\; \mathcal{L}_i^d
  + \sum_{b \in \mathcal{I}^{Const}}\sum_{a \in A}\sum_{\substack{p \in \mathcal{P}^d \\ a \in p}}
    \beta_b^a\, b^d\, s_{i}^p(b)
  - \gamma^d,
\end{equation}
where $\beta_b^a \geq 0$ is the dual of the capacity constraint for arc
$a$ and demand vector $b$, and $\gamma^d$ is the dual of the convex
combination constraint for demand $d$. The pricing problem minimizes
$r_i^d$ over the network weights $w_i^d$.

\paragraph{Cutting.}
The set $\mathcal{I}^{Const}$ may be infinite in size. The separation problem searches for a demand vector $b \in \mathcal{I}^{Const} \setminus \mathcal{I}^{RMP}$
that maximizes the utilization violation on any arc $a \in A$:
\begin{equation}\label{eq:sep-mcf}
\max \;
  \sum_{d \in D} b^d
  \sum_{i \in \mathcal{M}^d}  \sum_{\substack{p \in \mathcal{P}^d \\ a \in p}}s_{i}^p(b)\,\lambda_i^d \;-\; c_a.
\end{equation}
This is solved by a black-box optimizer (NLopt), with one thread per arc
to allow parallelization. The most violating demand vector $b^*$ and arc $a^*$ are added in $\mathcal{I}^{RMP}$ as a new capacity constraint.

\subsection{Numerical Results}\label{sub:mcf-results}

We evaluate CG4AI on the MCFP using benchmark instances from
SNDLIB~\citep{sndlib}, a standard library for network design problems.
SNDLIB provides both real-world and synthetic telecommunication
topologies, making it a widely used benchmark for routing algorithms.

\paragraph{Experimental setup.}

The number of demands per instance is chosen from $\{5, 10, 15\}$.
Candidate paths per demand are computed using Yen's
algorithm~\citep{yen1970algorithm} to obtain the $K$-shortest paths;
each demand has between 3 and 5 candidate paths. The training set
consists of 1000 randomly generated demand vectors with LP-optimal
routing splits (solved by CPLEX). Models are validated on 10\,000
random demand vectors.

\subparagraph{Architecture.} All column models are feedforward networks trained in PyTorch. A
hyperparameter search over hidden layer sizes $\{64, 128, 256\}$ and
depths $\{1, 2\}$ was conducted; the best configuration uses a single
hidden layer with 256 neurons and GELU activation. Training uses
Adam~\citep{kingma2014adam} with learning rate $10^{-3}$, early stopping
(patience 50), and a learning rate decay of $0.9$ every 10 epochs without
improvement (up to 1000 epochs). The LP is solved by CPLEX on an Intel
Xeon Platinum 8164; the separation problem is parallelized across arcs
(one thread per arc).

\paragraph{Results.}

Table~\ref{tab:results} (Appendix) reports detailed results for all
instances. Here we highlight the main findings.

\begin{figure}[ht]
    \centering
    \begin{subfigure}[b]{0.48\textwidth}
        \includegraphics[width=\textwidth]{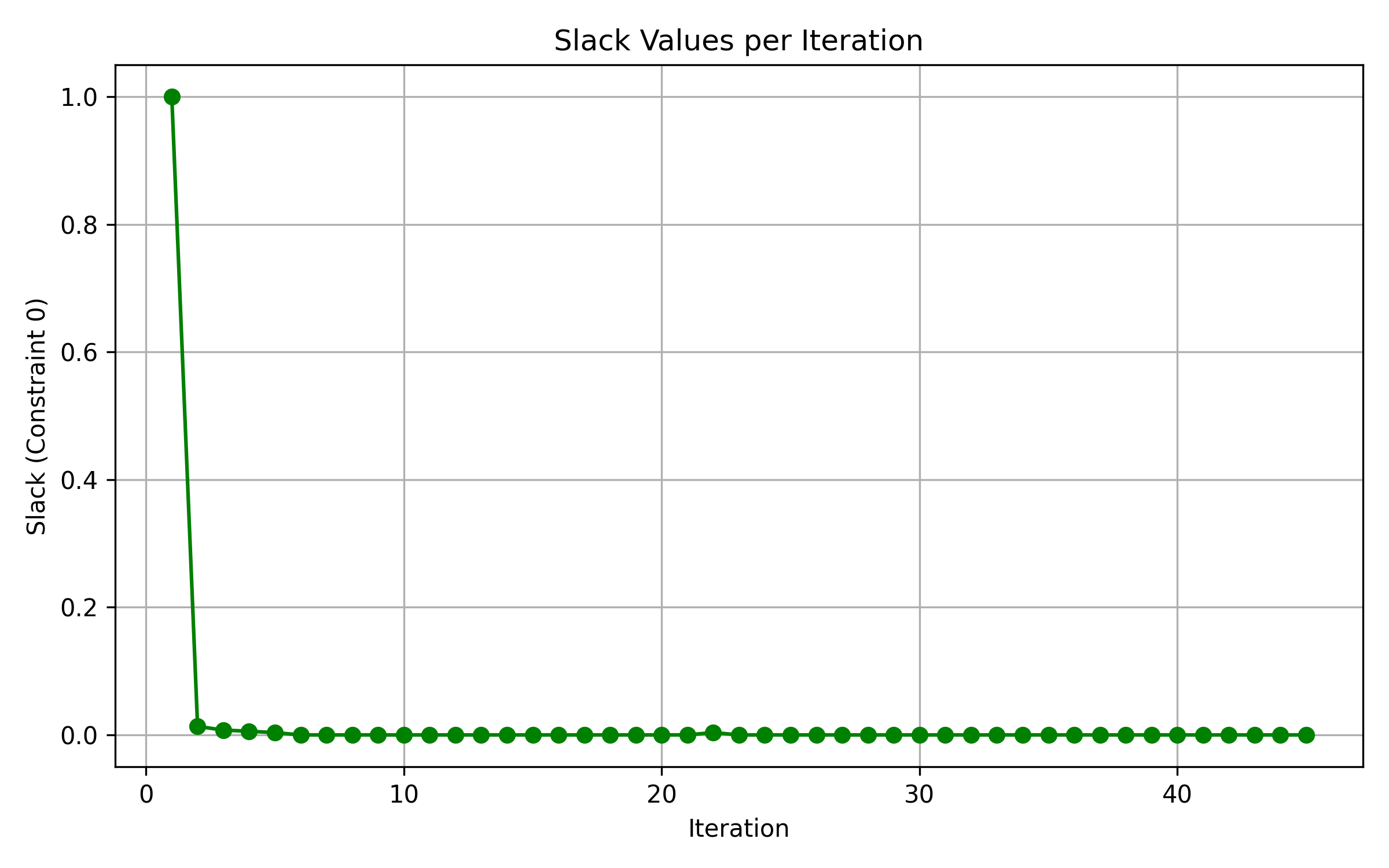}
        \caption{Slack evolution for nobel-us
                 $(\#d, \#p) = (5, 3)$.}
        \label{fig:nobel-us}
    \end{subfigure}
    \hfill
    \begin{subfigure}[b]{0.48\textwidth}
        \includegraphics[width=\textwidth]{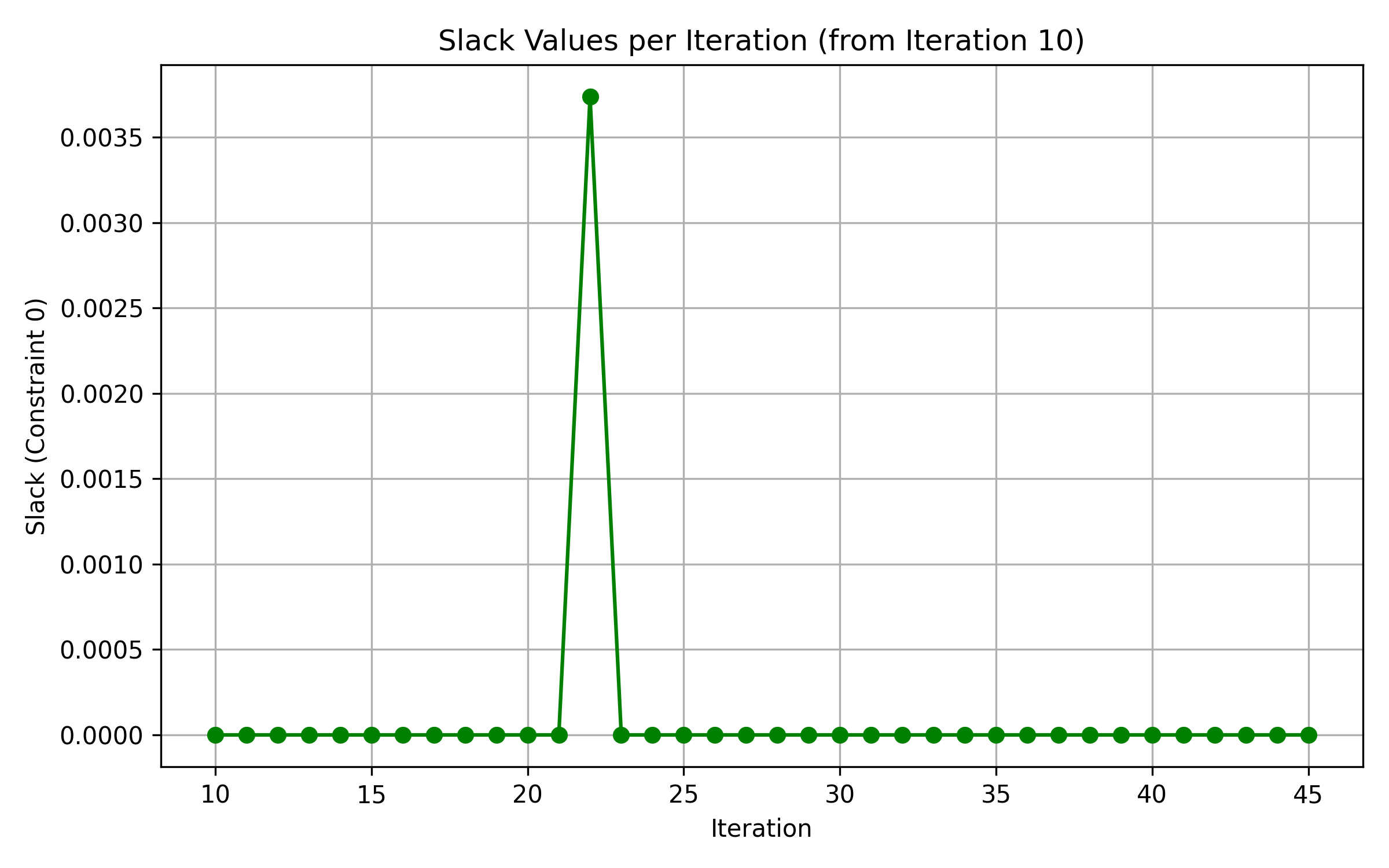}
        \caption{Zoomed view from iteration 10.\\~}
        \label{fig:slack_zoom}
    \end{subfigure}
    \caption{Slack variable evolution during the column-generation and
             cutting-plane iterations. The slack converges rapidly to
             zero, confirming that the algorithm achieves feasibility.}
    \label{fig:slack_side_by_side}
\end{figure}

\paragraph{Feasibility.}
CG4AI successfully produces feasible ensembles (zero slack) for most
instances, as shown by the slack convergence plots in
Figure~\ref{fig:slack_side_by_side}. The slack variable converges
rapidly to zero, with occasional small increases when a new constraint is
added, which is quickly resolved by the next column generation step.

\begin{figure}[ht]
    \centering
    \includegraphics[width=0.8\textwidth]{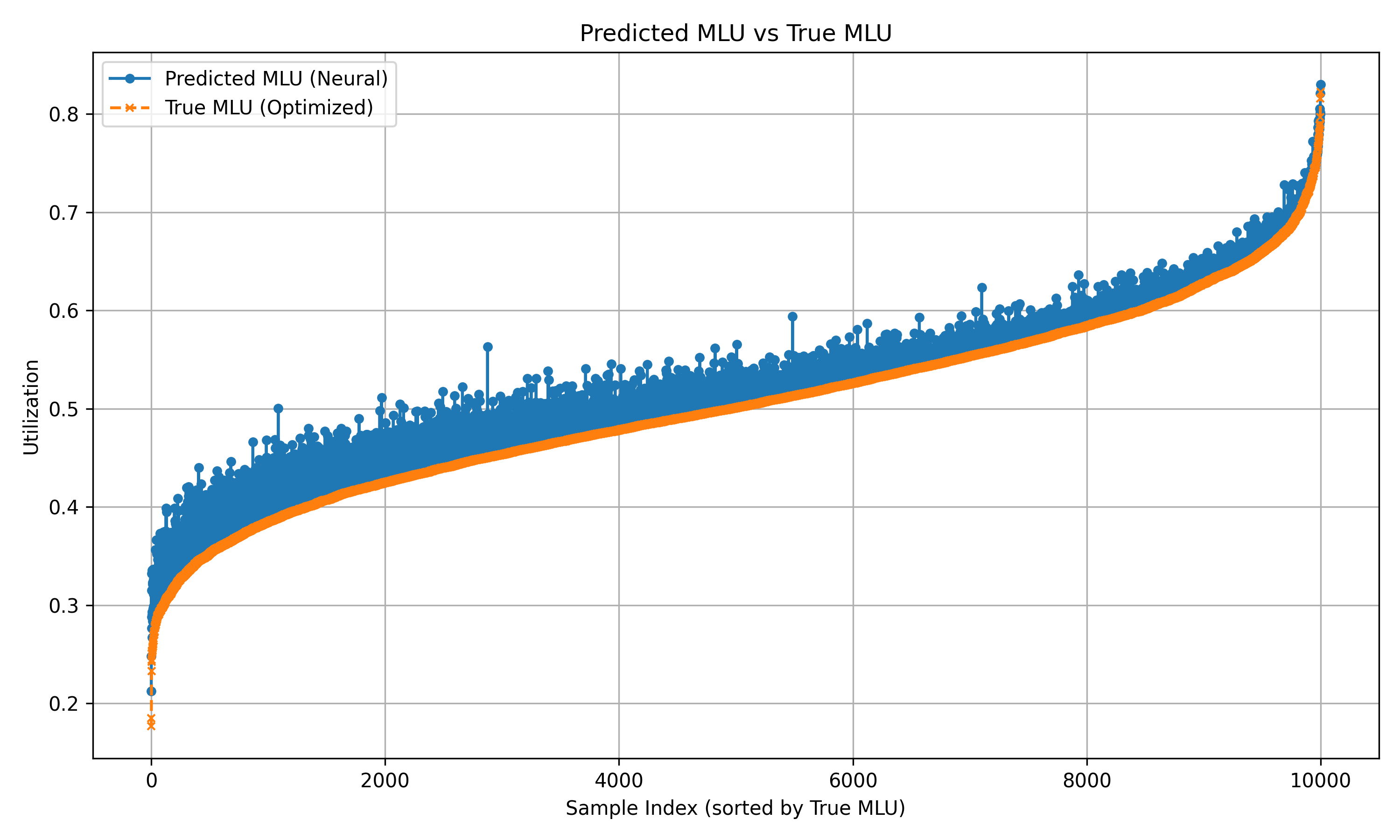}
    \caption{Predicted vs.\ optimal MLU on the nobel-germany instance
             $(\#d,\#p) = (15,3)$. The ensemble closely tracks the
             optimal solution.}
    \label{fig:nobel-germany}
\end{figure}

\begin{figure}[ht]
    \centering
    \includegraphics[width=0.8\textwidth]{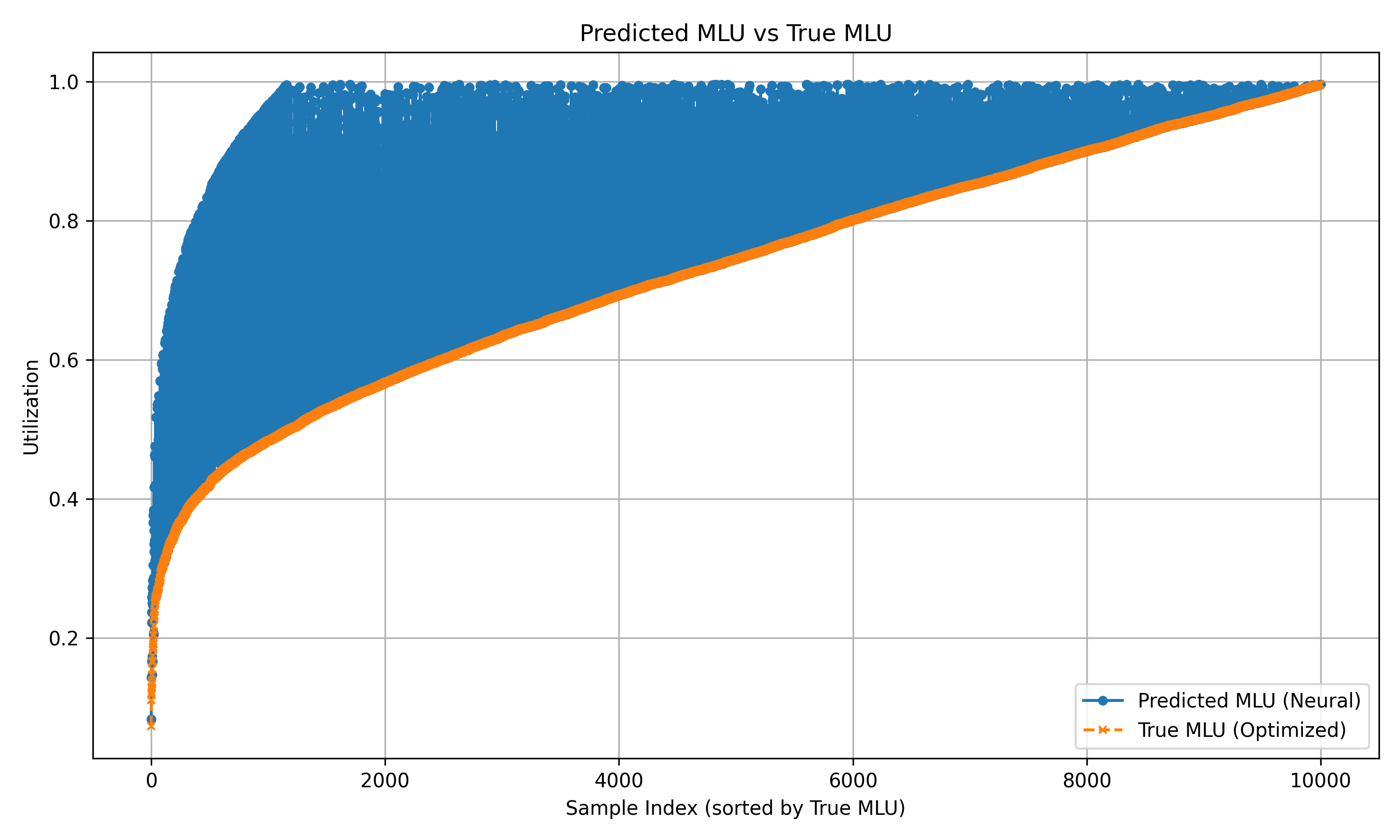}
    \caption{Predicted vs.\ optimal MLU on the newyork instance
             $(\#d,\#p) = (5,3)$. Greater dispersion occurs for
             high-utilization scenarios, but predictions remain feasible.}
    \label{fig:newyork}
\end{figure}

\paragraph{Prediction accuracy.}
Figure~\ref{fig:nobel-germany} shows the predicted versus optimal MLU
for the nobel-germany instance with 15 demands and 3 paths. The predicted
values align well with the optimal values. Figure~\ref{fig:newyork}
shows the newyork instance (5 demands, 3 paths), where predictions are
less accurate for high-utilization scenarios; the neural ensemble tends
to underestimate MLU in the most congested cases. Importantly, all
predictions remain below the critical value of 1.0 for both instances,
confirming constraint satisfaction.

\begin{figure}[ht]
    \centering
    \begin{subfigure}[b]{0.48\textwidth}
        \includegraphics[width=\textwidth]{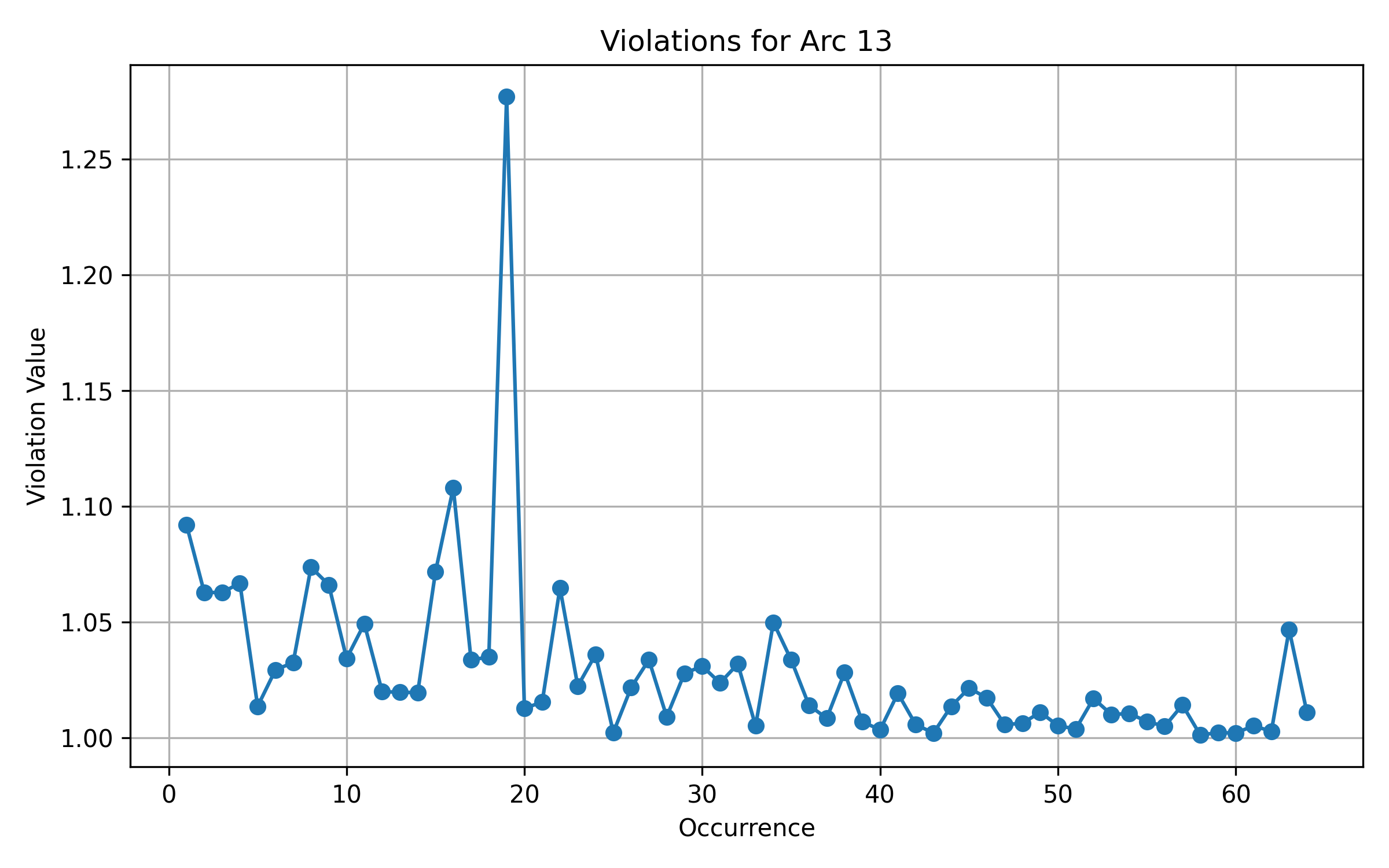}
        \caption{Utilization of Arc~13 over iterations.}
        \label{fig:arc13-viol}
    \end{subfigure}
    \hfill
    \begin{subfigure}[b]{0.48\textwidth}
        \includegraphics[width=\textwidth]{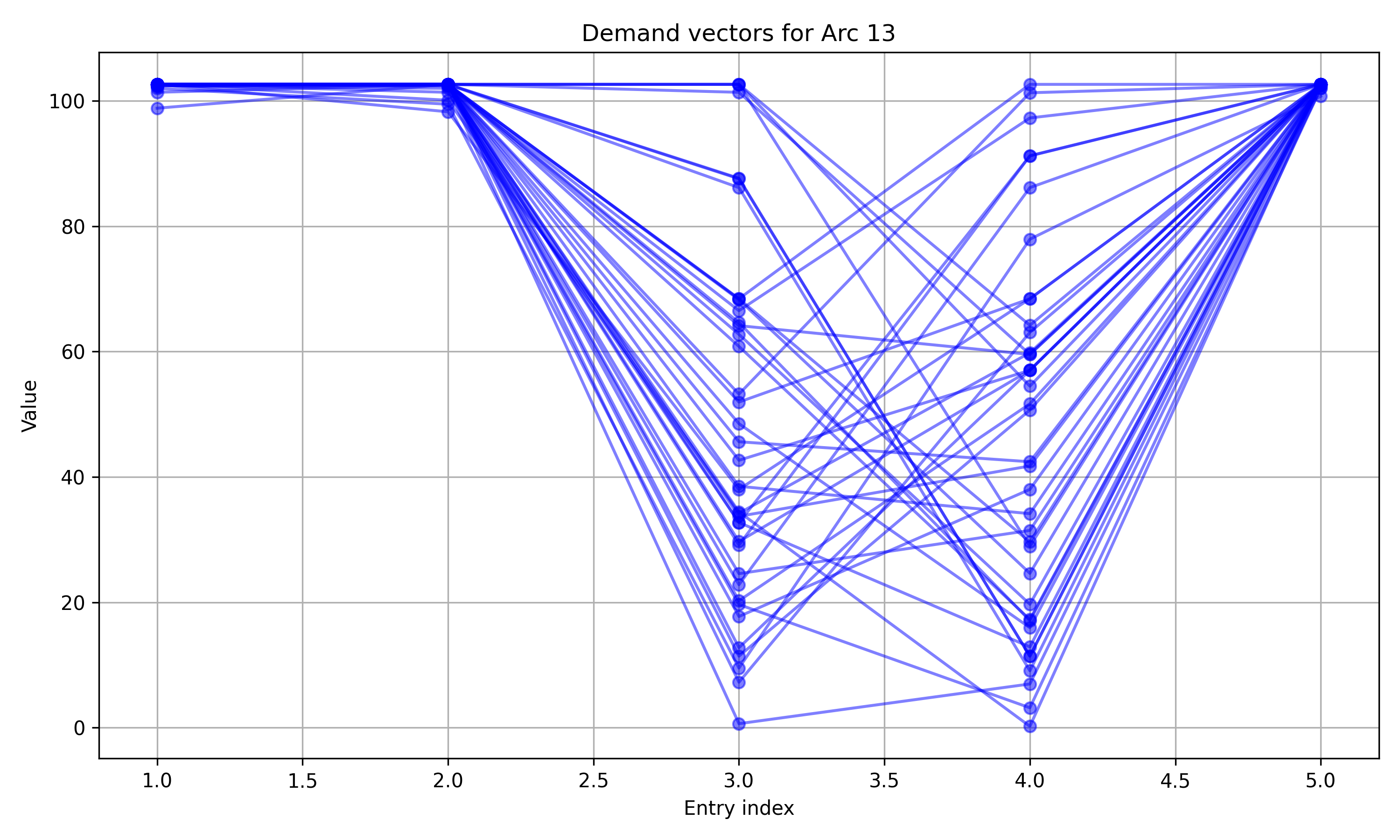}
        \caption{Demand vectors generated for Arc~13.}
        \label{fig:arc13-dem}
    \end{subfigure}
    \caption{Arc violations for Arc~13 in the Polska instance
             $(\#d, \#p) = (5, 3)$.}
    \label{fig:violation_arc_13}
\end{figure}

\paragraph{Constraint analysis.}
Figure~\ref{fig:violation_arc_13} shows the violation history for Arc~13
in the Polska instance during optimization process. Initial violations exceed the capacity by 25\%.
The violations decrease over iterations, though not monotonically,
reflecting the interplay between adding new columns and new constraints.

The analysis of the demand vectors generated for Arc~13 reveals an
informative pattern: three demands are consistently set to their maximum
value $m$, while the remaining two are varied across $[0,m]$. Inspection
of the network topology shows that those three demands have no path
avoiding Arc~13, making them \emph{captive} to this bottleneck. The
optimizer varies the non-captive demands to find combinations where the
ensemble's routing fails to exploit alternative paths. This insight could
be used to seed the initial constraint set $\mathcal{I}^{Const}_0$ with
adversarial scenarios (captive demands at $m$, others sampled
strategically), potentially reducing the number of cutting-plane
iterations.

\paragraph{Instance-level analysis.}
Across the SNDLIB benchmark (Table~\ref{tab:results}), the results
confirm several consistent patterns. The algorithm produces feasible
solutions for most instances. Harder instances (more demands, denser
topologies) require more iterations and generate more columns; for
example, di-yuan with 10 demands requires 999 iterations and 986 columns.
Simpler instances converge in just 1--3 iterations (ta1 with 5 demands,
france with 10 demands). The nobel family (eu, germany, us) consistently
achieves very precise predictions, with validation losses on the order
of $10^{-4}$. The final active model count is typically much smaller than
the total number of generated columns, confirming that the LP selects a
compact ensemble. For three instances (geant with 15 demands, di-yuan with
15 demands and 10 demands), the iteration limit is reached without full convergence,
suggesting that these configurations require a larger column budget.

\paragraph{Ablation: alternative objective.}

To verify that constraint satisfaction in CG4AI is driven by the
dual-guided pricing and not by properties of the MLU objective itself, we
replace the MLU with an alternative cost function: the total weighted
path cost $\sum_{d \in D}\sum_{p \in \mathcal{P}^d} w_p \cdot f_p^d
\cdot b^d$, where $w_p$ is the sum of arc weights along path $p$. This
function has no inherent regularizing effect on capacity constraints.\\

\begin{figure}[H]
    \centering
    \begin{subfigure}[b]{0.7\textwidth}
        \includegraphics[width=\textwidth]{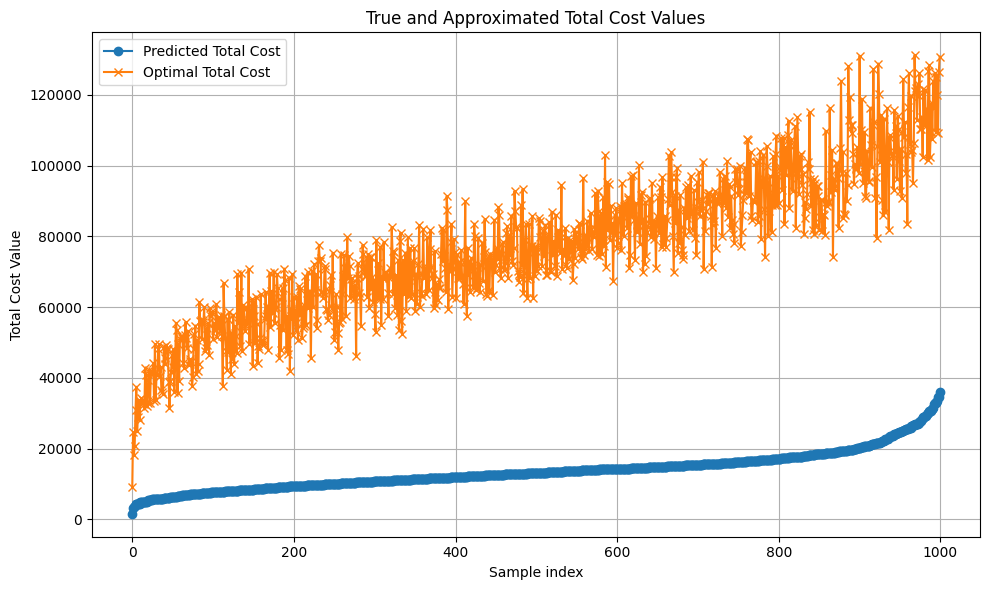}
        \caption{Predicted vs.\ optimal total path cost.}
    \end{subfigure}
    \hfill
    \begin{subfigure}[b]{0.7\textwidth}
        \includegraphics[width=\textwidth]{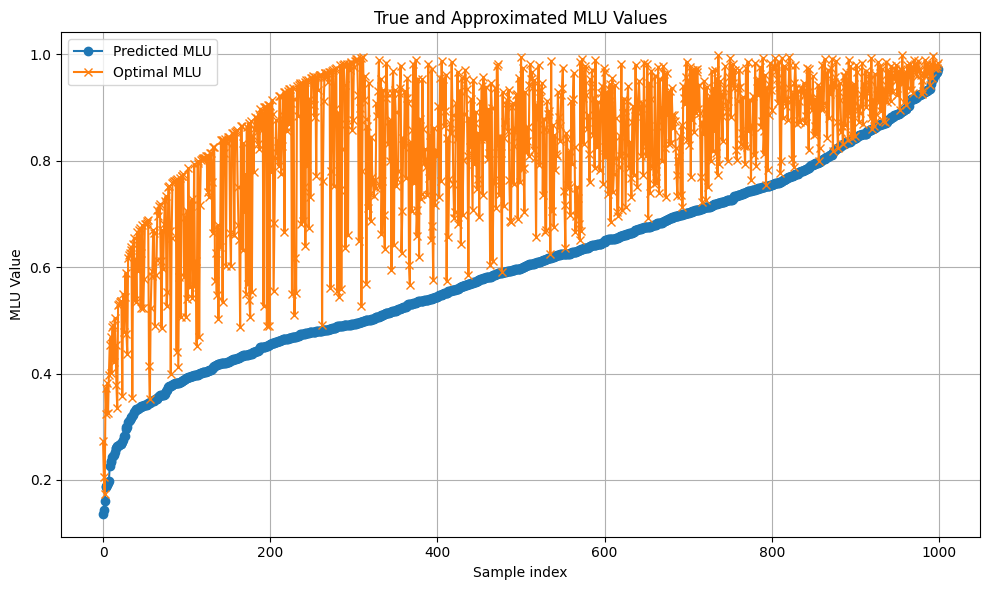}
        \caption{MLU while optimizing the total path cost.}
    \end{subfigure}
    \caption{Constrained optimization with the alternative cost function
             on the ta1 instance. CG4AI satisfies all capacity constraints
             even though the training objective is not aligned with them.}
    \label{fig:constrained_optimization}
\end{figure}

Figure~\ref{fig:constrained_optimization} shows the results on the ta1
instance with 10 demands and 2 paths. CG4AI still produces a feasible
ensemble that respects all capacity constraints, although more iterations
are needed compared to the MLU case. The optimality gap on the cost
function is larger (11.20\% at the final iteration), reflecting the
harder learning problem when the objective is less aligned with the
constraints. This confirms that the dual-guided pricing is the primary
mechanism for enforcing constraint satisfaction, independent of the
specific training objective.

\section{Conclusion}\label{sec:conclusion}

We have presented CG4AI, a column generation framework for training
convex ensembles of AI models that satisfy hard linear constraints. The
framework combines three key ingredients: a master LP that enforces
feasibility through the mixture weights; a pricing subproblem that trains
new models guided by LP dual variables to focus on the most violated
constraints; and a cutting-plane procedure that extends feasibility
guarantees beyond a finite training set.

We demonstrated CG4AI on two applications. On MNIST, we showed that hard
constraints can be used as a replacement for labeled data (constraint-only
learning), as a mechanism for improving adversarial robustness, as a tool
for correcting misclassified examples, and as a way to implement output
relabeling without full retraining. On the multi-commodity flow problem,
CG4AI produces routing predictors that satisfy all link capacity
constraints on standard SNDLIB benchmark networks, with good MLU accuracy
and fast inference at test time.

The results highlight several interesting properties of the framework.
The \texttt{margin} constraint formulation is particularly efficient,
requiring only 3--6 columns even for 1000 training images, with zero
residual violation. The dual-guided pricing is effective regardless of
the training objective, as shown by the ablation with an alternative cost
function. The cutting-plane separation provides a principled mechanism
for finding adversarial demand vectors that expose model weaknesses.

Several directions remain open for future work. First, enforcing an integer ensemble by restricting $\lambda_i \in \{0, 1\}$ via a branch-and-bound scheme would yield a single, highly interpretable model. Second, the model-agnostic nature of the master LP framework permits the integration of alternative column classes, such as decision trees or gradient-boosted models, which are well-suited for structured constraint patterns. Third, computational efficiency could be significantly enhanced by replacing the general black-box separator with problem-specific cut generation heuristics, or by employing an exact ILP-based separation formulation for single-layer ReLU networks to guarantee provably optimal cuts. Finally, the framework can be extended dynamically through an active learning procedure that updates training losses using newly identified violating inputs, or scaled to distributed AI agents to enable decentralized deployment while maintaining global feasibility guarantees.

\bibliographystyle{plainnat}
\bibliography{refs}

\newpage

\begin{appendices}

\section{Experimental Tables}\label{app:tables}

\begin{table}[ht]
\centering
\caption{Column definitions for Table~\ref{tab:results}.}\label{tab:params}
\begin{tabularx}{\textwidth}{lX}
\toprule
Column & Description \\
\midrule
instance   & Network instance name. \\
\#d        & Number of demands. \\
\#p        & Number of candidate paths per demand. \\
\#iter     & Total iterations of the combined column-generation and cutting-plane procedure. \\
\#cols     & Total columns generated. \\
\#rows     & Constraints (rows) added by the cutting-plane step. \\
col\_time  & Time spent on pricing (training AI models), in seconds. \\
row\_time  & Time spent by the black-box separator, in seconds. \\
tot\_time  & Total computation time, in seconds. \\
opt\_mlu   & Optimal MLU of the LP (ground truth). \\
\#mod      & Number of active columns in the final ensemble (non-zero $y$ coefficients). \\
val.\ loss & Validation loss (aggregated weighted loss on 10\,000 unseen demand vectors). \\
max\_diff  & Maximum absolute difference between predicted and optimal MLU on validation set. \\
\bottomrule
\end{tabularx}
\end{table}

\begin{footnotesize}
\setlength{\LTleft}{-3cm}
\setlength{\LTright}{0cm}
\begin{longtable}{lrrrrrrrrrrrr}
\caption{Summary of CG4AI results on SNDLIB instances.}
\label{tab:results} \\
\toprule
instance & \#d & \#p & \#iter & \#cols & \#rows & col\_time & row\_time & tot\_time & opt\_mlu & \#mod & val.\ loss & max\_diff \\
\midrule
\endfirsthead
\multicolumn{13}{l}{\small\textit{(continued from previous page)}} \\
\toprule
instance & \#d & \#p & \#iter & \#cols & \#rows & col\_time & row\_time & tot\_time & opt\_mlu & \#mod & val.\ loss & max\_diff \\
\midrule
\endhead
\bottomrule
\endfoot
abilene        & 5  & 3 &   4 &   4 &  0 &    284.83 &    875.47 &   1160.41 & 7.63e-02 &   2 & 0.021 & 0.38 \\
               & 10 & 3 &  57 &  36 &  9 &   2207.11 &   7831.87 &  10040.36 & 1.57e-01 &  34 & 0.013 & 0.46 \\
               & 15 & 3 &  28 &  21 &  0 &   1336.92 &   1215.77 &   2553.48 & 4.26e-02 &   3 & 0.015 & 0.42 \\
               &  5 & 5 &  11 &   8 &  4 &    463.05 &   1190.41 &   1653.69 & 9.14e-02 &  10 & 0.027 & 0.42 \\
atlanta        &  5 & 3 &  45 &  26 & 10 &   2229.10 &  13427.90 &  15658.01 & 6.07e-01 &  43 & 0.010 & 0.33 \\
               & 10 & 3 &  63 &  42 & 16 &   2832.30 &  19045.30 &  21879.00 & 9.91e-02 &  53 & 0.017 & 0.37 \\
               & 15 & 3 & 127 &  59 &  5 &   6319.78 &  10198.40 &  16520.90 & 2.95e-02 &  26 & 0.027 & 0.57 \\
               &  5 & 5 &   5 &   5 &  0 &    388.27 &   1279.12 &   1667.53 & 5.16e-01 &   3 & 0.004 & 0.34 \\
dfn-bwin       &  5 & 3 &  28 &  24 &  0 &   3582.93 &    774.63 &   4358.14 & 1.01e-01 &   1 & 0.101 & 0.65 \\
               & 10 & 3 & 353 & 156 & 42 &  25530.80 &  95600.80 & 121141.19 & 2.90e-01 & 104 & 0.037 & 0.49 \\
               & 15 & 3 & 406 & 138 &  8 &  30845.70 &  22207.00 &  53064.82 & 2.73e-01 &  39 & 0.036 & 0.45 \\
di-yuan        &  5 & 3 & 155 & 147 &  8 &  12730.70 &  12305.70 &  25039.90 & 3.59e-03 &  33 & 0.003 & 0.16 \\
               & 10 & 3 & 999 & 986 & 59 &  71904.80 &  52162.70 & 124094.28 & 9.15e-03 & 143 & 0.009 & 0.23 \\
               & 15 & 3 & 999 & 845 &  0 &  68495.30 &      0.00 &  68516.25 & 6.09e+00 &  58 & 0.058 & 0.50 \\
               &  5 & 5 & 350 & 344 &  9 &  21822.80 &  14763.50 &  36594.96 & 4.45e-03 &  44 & 0.004 & 0.17 \\
france         & 10 & 3 &   3 &   3 &  0 &    273.24 &    241.05 &    514.36 & 2.28e-02 &   1 & 0.024 & 0.49 \\
               & 15 & 3 &   9 &   7 &  0 &   1300.78 &   1288.17 &   2589.15 & 2.88e-02 &   1 & 0.029 & 0.52 \\
               &  5 & 5 &   2 &   2 &  0 &    384.32 &    940.69 &   1325.07 & 5.97e-02 &   1 & NaN   & NaN  \\
geant          &  5 & 3 &   2 &   2 &  0 &    393.92 &   1058.57 &   1452.53 & 9.98e-02 &   1 & 0.099 & 0.66 \\
               & 10 & 3 &  16 &  14 &  0 &   1682.60 &   4160.63 &   5843.64 & 1.68e-01 &   7 & 0.040 & 0.48 \\
               & 15 & 3 & 999 & 298 & 13 &  55938.10 &  13270.10 &  69230.61 & 4.00e+02 &  87 & 0.042 & 0.45 \\
               &  5 & 5 &  23 &  17 &  4 &   1683.58 &   2526.60 &   4210.79 & 1.08e-01 &  16 & 0.102 & 0.66 \\
newyork        &  5 & 3 &  12 &  11 &  0 &   1969.16 &    993.96 &   2963.48 & 2.39e-02 &   1 & 0.025 & 0.49 \\
               & 10 & 3 &  60 &  29 & 22 &   4803.46 &  49169.50 &  53974.74 & 1.12e-01 &  71 & 0.017 & 0.34 \\
               & 15 & 3 &  20 &  14 &  2 &   2502.16 &   5126.39 &   7629.03 & 5.20e-02 &   9 & 0.049 & 0.56 \\
               &  5 & 5 &   1 &   1 &  0 &    288.41 &   1080.68 &   1369.11 & 5.90e-02 &   1 & 0.056 & 0.61 \\
nobel-eu       &  5 & 3 & 886 & 868 & 24 &  56205.70 &  14245.10 &  70476.87 & -5.37e-02 & 66 & 0.001 & 0.13 \\
               & 10 & 3 &  63 &  45 & 17 &   5047.61 &  31214.60 &  36263.94 & 1.98e-04 &  57 & 2.3e-4 & 0.10 \\
               & 15 & 3 & 209 & 180 & 29 &  15888.80 &  45788.20 &  61682.89 & 3.64e-04 &  77 & 6.0e-4 & 0.13 \\
nobel-germany  &  5 & 3 & 180 & 131 & 94 &   7834.62 &  45883.70 &  53722.76 & 3.40e-04 & 189 & 3.2e-4 & 0.09 \\
               & 15 & 3 &  38 &  32 &  4 &   2753.76 &   5227.27 &   7982.08 & 2.47e-04 &  17 & 4.1e-4 & 0.11 \\
               &  5 & 5 & 142 &  96 & 79 &   5303.81 &  37377.30 &  42684.53 & 1.87e-04 & 164 & 1.3e-4 & 0.07 \\
nobel-us       &  5 & 3 &  46 &  34 &  8 &   2605.42 &   7542.89 &  10149.66 & 5.81e-05 &  30 & 7.1e-5 & 0.05 \\
               & 10 & 3 & 258 & 167 & 75 &   8832.34 &  85008.30 &  93847.17 & 4.18e-04 & 234 & 6.0e-4 & 0.14 \\
               & 15 & 3 & 189 & 167 & 47 &   8804.43 &  33584.50 &  42393.73 & 1.01e-03 & 100 & 0.001  & 0.15 \\
               &  5 & 5 &  19 &  15 &  3 &    974.58 &   2233.84 &   3208.82 & 1.03e-04 &  13 & 1.1e-4 & 0.06 \\
polska         &  5 & 3 & 211 & 115 & 128 &  5319.47 &  55888.30 &  61212.43 & 4.91e-03 & 282 & 0.002  & 0.22 \\
ta1            &  5 & 3 &   1 &   1 &  0 &    386.52 &   1291.10 &   1677.64 & 2.01e-02 &   1 & 0.020  & 0.49 \\
               & 10 & 3 & 142 &  65 & 58 &   9850.20 & 151503.00 & 161356.94 & 7.31e-02 & 175 & 0.018  & 0.42 \\
               & 15 & 3 &  72 &  27 &  0 &   9054.51 &   4745.62 &  13802.14 & 7.07e-02 &   6 & 0.021  & 0.42 \\
               &  5 & 5 &   4 &   4 &  0 &    730.99 &   1773.09 &   2504.20 & 4.75e-02 &   2 & 0.047  & 0.66 \\
\end{longtable}
\end{footnotesize}

\paragraph{Note on anomalous \texttt{opt\_mlu} values.}
\textit{geant~(15d)}: \texttt{opt\_mlu}$=4.00\times10^2$ — the LP becomes unbounded for this demand/capacity configuration (demand exceeds capacity); CG4AI still generates a feasible ensemble within the path set. \textit{nobel-eu~(5d)}: \texttt{opt\_mlu}$=-5.37\times10^{-2}$ — floating-point artefact of the CPLEX dual at near-zero MLU; functionally equivalent to $0$.

The results across SNDLIB instances confirm the following patterns. The
algorithm produces feasible solutions for the large majority of tested
configurations. Computational effort scales with problem difficulty:
di-yuan and polska are the hardest instances, requiring hundreds to
nearly a thousand iterations, while france and ta1 converge in just a
few. The nobel family achieves the highest accuracy, with validation
losses on the order of $10^{-4}$ and maximum differences below 0.15 MLU
units. The final active model count (column \#mod) is consistently
smaller than the total number of generated columns, indicating that the
LP selects a compact ensemble that covers the constraint requirements
without retaining all generated columns.

\end{appendices}

\end{document}